\documentclass[twoside]{article} \usepackage{aistats2017}

\usepackage[square,sort,comma,numbers]{natbib}%

\usepackage[utf8]{inputenc} 
\usepackage[T1]{fontenc}    
\usepackage{hyperref}       
\usepackage{url}            
\usepackage{booktabs}       
\usepackage{amsfonts}       
\usepackage{nicefrac}       
\usepackage{microtype}      

\usepackage{soul}
\usepackage{epsf,psfrag,amssymb,amsfonts,latexsym,graphicx,bm,xcolor,url,array}
\usepackage{subcaption}
\usepackage{float}
\usepackage{fixltx2e}
\usepackage{verbatim}
\usepackage[mathscr]{eucal}
\usepackage[tbtags]{mathtools}
\usepackage{multirow}
\usepackage{amsmath}
\usepackage{thmtools,thm-restate}
\usepackage{algorithm}
\usepackage{algpascal}
\usepackage{algc}
\usepackage{algcompatible}
\usepackage{algpseudocode}
\usepackage[utf8]{inputenc}
\usepackage[english]{babel}
\usepackage{authblk}
\usepackage{wrapfig}
\usepackage{tabularx}

\usepackage{tikz}
\usetikzlibrary{positioning,arrows,shapes}

\tikzstyle{controller}=[circle,draw=blue!50,fill=blue!30,thick,
inner sep=0pt,minimum size=20mm]

\tikzstyle{state} = [circle,draw=black!100,fill=blue!20,thick,
inner sep=0pt,minimum size=20mm
]

\tikzstyle{observ}=[circle,draw=black!100,fill=black!10,thick,
inner sep=0pt,minimum size=20mm]

\tikzstyle{pre}=[<-,shorten <=1pt,>=stealth',thick]

\tikzstyle{post}=[->,shorten >=1pt,>=stealth',thick]

\tikzstyle{block} = [rectangle, draw, fill=blue!0
,text width=25em
, text centered, rounded corners
, minimum height=2em
]

\tikzstyle{blockred} = [rectangle, draw, fill=blue!30
,text width=25em
, text centered, rounded corners
, minimum height=2em
]

\tikzstyle{blockS} = [rectangle, 
,text width=8em
, text centered, rounded corners
, minimum height=2em
]

\tikzstyle{block2} = [rectangle, draw, fill=blue!0
,text width=8em
, text centered, rounded corners
, minimum height=2em
]
\tikzstyle{block2red} = [rectangle, draw, fill=red!0
,text width=8em
, text centered, rounded corners
, minimum height=2em
]
\tikzstyle{line} = [draw, -latex']

\tikzstyle{cloud} = [draw, ellipse
]

\tikzstyle{blockblue} = [rectangle, draw, fill=red!0
,text width=25em
, text centered, rounded corners
, minimum height=3em
]
\tikzstyle{blockred} = [rectangle, draw, fill=red!0
,text width=25em
, text centered, rounded corners
, minimum height=3em
]
\tikzstyle{blockyel} = [rectangle, draw, fill=blue!15
,text width=15em
, text centered, rounded corners
, minimum height=3em
]
\tikzstyle{line} = [draw, -latex']
\tikzstyle{cloud} = [draw, ellipse,fill=orange!0
]

\tikzstyle{blockeqn} = [rectangle, draw, fill=blue!15
,minimum width=5em
, text centered, rounded corners
, minimum height=1em
]

\newcommand{\xv}{{\mathbf x}}

\newcommand{\hv}{{\mathbf h}}
\newcommand{\Am}{{\mathbf A}}

\usepackage[explicit]{titlesec}
\titlespacing{\paragraph}{0pt}{0.5em}{0.5em}[]

\def\tha{{\mbox{\tiny th}}}

 \def\0{{\bf 0}}

\def\qed{\hfill\hbox{${\vcenter{\vbox{
    \hrule height 0.4pt\hbox{\vrule width 0.4pt height 6pt
    \kern5pt\vrule width 0.4pt}\hrule height 0.4pt}}}$}}

\definecolor{myred}{rgb}{0.3,0.0,0.7}
\definecolor{dkg}{rgb}{0.1,0.7,0.2}
\definecolor{dkb}{rgb}{0.0,0.2,0.8}

\def\bfa{{\mathbf a}}

\def\bfe{{\mathbf e}}

\def\bfx{{\mathbf x}}

\def\bfA{{\mathbf A}}

\def\bfM{{\mathbf M}}

\def\alphabf{\hbox{\boldmath$\alpha$\unboldmath}}

\def\Ebb{{\mathbb E}}

\def\Rbb{{\mathbb R}}

\newcommand{\bprf}{\begin{myproof}}
\newcommand{\eprf}{\end{myproof}}
\newcommand{\bp}{\begin{psfrags}}
\newcommand{\ep}{\end{psfrags}}
\newcommand{\bl}{\begin{lemma}}
\newcommand{\el}{\end{lemma}}
\newcommand{\bt}{\begin{theorem}}
\newcommand{\et}{\end{theorem}}
\newcommand{\bc}{\begin{center}}
\newcommand{\ec}{\end{center}}
\newcommand{\bi}{\begin{itemize}}
\newcommand{\ei}{\end{itemize}}
\newcommand{\ben}{\begin{enumerate}}
\newcommand{\een}{\end{enumerate}}
\newcommand{\bd}{\begin{definition}}
\newcommand{\ed}{\end{definition}}
\def\beq{\begin{equation}}
\def\eeq{\end{equation}\noindent}
\def\beqn{\begin{eqnarray}}
\def\eeqn{\end{eqnarray} \noindent}
\def\beqnn{  \begin{eqnarray*}}
\def\eeqnn{\end{eqnarray*}  \noindent}
\def\bcase{  \begin{numcases}}
\def\ecase{\end{numcases}   \noindent}
\def\bsbcase{  \begin{subnumcases}}
\def\esbcase{\end{subnumcases}   \noindent}

\newtheorem{theorem}{Theorem}
\newtheorem{corollary}{Corollary}
\newtheorem{lemma}{Lemma}

\newtheorem{assumption}{Assumption}

\newenvironment{myproof}{\noindent{\em Proof:} \hspace*{1em}}{
    \hspace*{\fill} $\Box$ }
\newenvironment{proof_of}[1]{\noindent {\em Proof of #1: }}{\hspace*{\fill} $\Box$ }

\newcommand{\matplottc}[1]{               
        \unitlength .45truein
        \begin{center}
        \includegraphics{#1.ps}
        \end{picture}
        \end{center}
}

\def\psfancypar#1#2{\begingroup\def\par{\endgraf\endgroup\lineskiplimit=0pt}
               \setbox2=\hbox{\large\sc #2}
               \newdimen\tmpht \tmpht \ht2 \advance\tmpht by \baselineskip
               \font\hhuge=Times-Bold at \tmpht
               \setbox1=\hbox{{\hhuge #1}}
               \count7=\tmpht \count8=\ht1
               \divide\count8 by 1000 \divide\count7 by \count8
               \tmpht=.001\tmpht\multiply\tmpht by \count7
               \font\hhuge=Times-Bold at \tmpht
               \setbox1=\hbox{{\hhuge #1}}
               \noindent
                \hangindent1.05\wd1
               \hangafter=-2 {\hskip-\hangindent
               \lower1\ht1\hbox{\raise1.0\ht2\copy1}%
                \kern-0\wd1}\copy2\lineskiplimit=-1000pt}

\def\Kout{\setbox1=\hbox{\Huge\bf K}\hbox to
1.05\wd1{\hspace{.05\wd1}
}}
\def\Sout{\setbox1=\hbox{\Huge\bf S}\hbox to 1.05\wd1{\hspace{.05\wd1}
}}

\allowdisplaybreaks[4]

\begin{document}

%

%

\twocolumn[

\aistatstitle{Spectral Methods for Correlated Topic Models}

\aistatsauthor{ Forough Arabshahi$^\dagger$ \And Animashree Anandkumar }

\aistatsaddress{ University of California Irvine \And University of California Irvine} ]

\begin{abstract}
In this paper we propose guaranteed spectral methods for learning a broad range of topic models, which generalize the popular Latent Dirichlet Allocation (LDA). We overcome the limitation of LDA to incorporate arbitrary topic correlations, by assuming that the hidden topic proportions are drawn from  a flexible class of Normalized Infinitely Divisible (NID) distributions. NID distributions are generated through the process of normalizing a family of independent Infinitely Divisible (ID) random variables. The Dirichlet distribution is a special case obtained by normalizing a set of Gamma random variables. We prove that this flexible topic model class can be learnt via spectral methods using only moments up to the third order, with (low order) polynomial sample and computational complexity. The proof is based on a key new technique derived here that allows us to diagonalize the moments of the NID distribution through an efficient procedure that requires evaluating only univariate integrals, despite the fact that we are handling high dimensional multivariate moments. In order to assess the performance of our proposed Latent NID topic model, we use two real datasets of articles collected from New York Times and Pubmed. Our experiments yield improved perplexity on both datasets compared with the baseline. 
\end{abstract}

\section{Introduction}


Topic models are a popular class of exchangeable latent variable models for document categorization. The goal is to uncover hidden topics based on the distribution of word occurrences in a document corpus.
Topic models are {\em admixture} models, which go beyond the usual mixture model that allows for only one hidden topic to be present in each document. In contrast, topic models incorporate multiple topics in each document. It is assumed that each document has a latent proportions of different topics, and the observed words are drawn in a conditionally independent manner, given the set of topics.

Latent Dirichlet Allocation (LDA) is the most popular topic model~\cite{blei2003latent}, in which the topic proportions are drawn from the Dirichlet distribution. While LDA has widespread applications, it is limited by the choice of the Dirichlet distribution. Notably, Dirichlet distribution can only model negative correlations~\cite{bakhtiari2014online}, and thus, is unable to incorporate arbitrary correlations among the topics that may be present in different document corpora. Another drawback is that the elements with similar means need to have similar variances. While there have been previous attempt to go beyond the Dirichlet distribution, e.g.~\cite{blei2006correlated,sato2010topic}, their correlation structures are still limited,  learning these models is usually difficult and no guaranteed algorithms exist. Furthermore, As discussed in \cite{passos2011correlations}, the correlation structure considered in \cite{blei2006correlated}, gives rise to spurious correlations resulting in a better perplexity on the held-out set even when the recovered topics are less interpretable. The work of \cite{arora2013practical} provides a provably correct algorithm for learning topic models that also allow for certain correlations among the topics, however, it requires ``anchor word'' separability assumptions for the proof of correctness.

In this work, we consider a flexible class of topic models, and propose guaranteed and efficient algorithms for learning them. We employ the class of Normalized Infinitely Divisible (NID) distributions to model the topic proportions~\cite{favaro2011class,mangili2015new}. These are a class of distributions on the simplex, formed by normalizing a set of independent draws from a family of positive Infinitely Divisible (ID) distributions. The draws from an ID distribution can be represented as a sum of an arbitrary number of i.i.d. random variables. The concept of infinite divisibility was introduced in 1929 by Bruno de Finetti, and the most fundamental results were developed by Kolmogorov, L\'{e}vy and Khintchine in the 1930s. The idea of using normalized random probability measures with independent increments have also been used in the context of non-parametric models to go beyond the Dirichlet Process \cite{lijoi2010models}.

The Gamma distribution is an example of an ID distribution, and  the Dirichlet distribution is obtained by normalizing a set of independent draws from   Gamma distributions. We show that the class of NID topic models significantly generalize the LDA model: they can incorporate both positive and negative correlations among the topics and they involve additional parameters to vary the variance and higher order moments, while fixing the mean.

There are mainly three categories of algorithms for learning topic models, viz., variational inference~\cite{blei2003latent,blei2006correlated}, Gibbs sampling~\cite{griffiths2004finding,mimno2008gibbs,chen2013scalable}, and spectral methods~\cite{anandkumar2012spectral,tung2014spectral}. Among them, spectral methods have gained increasing prominence over the last few years, due to their efficiency and guaranteed learnability. In this paper, we develop novel spectral methods for learning latent NID topic models.

Spectral methods have previously been proposed for learning LDA~\cite{anandkumar2012spectral}, and in addition, other latent variable models such as Independent Component Analysis (ICA), Hidden Markov Models (HMM) and mixtures of ranking distributions~\cite{anandkumar2014tensor}. The idea is to learn the parameters based on spectral decomposition of low order moment tensors (third or fourth order). Efficient algorithms for tensor decomposition have been proposed before~\cite{anandkumar2014tensor}, and implies consistent learning with (low order) polynomial computational and sample complexity.

The main difficulty in extending spectral methods to the more general class of NID topic models is the presence of arbitrary correlations among the hidden topics which need to be ``untangled''. For instance, take the  case of a single topic model (i.e. each document has only one topic); here, the third order moment, which is the co-occurrence tensor of word triplets, has a CANDECOMP/PARAFAC (CP) decomposition, and computing the decomposition yields an estimate of the topic-word matrix. In contrast, for the LDA model, such a tensor decomposition is obtained by a combination of moments up to the third order. In other words, the moments of the LDA model need to be appropriately ``centered'' in order to have the tensor decomposition form.

Finding such a moment combination has so far been an ``art form'', since it is based on explicit manipulation of the moments of the hidden topic distribution. So far, there is no principled mechanism to automatically find the moment combination with the CP decomposition form.  For arbitrary topic models, however, finding such a combination may not even be possible. In general, one requires all the higher order moments for learning. 

In this work, we show that surprisingly, for the flexible class of NID topic models, moments up to third order suffice for learning, and we provide an efficient algorithm for computing the coefficients to combine the moments. The algorithm is based on computation of a univariate integral, that involves the Levy measure of the underlying ID distribution. The integral can be computed efficiently through numerical integration since it is only univariate, and has no dependence on the topic or word dimensions. Intriguingly, this can be accomplished, even when there exists no closed form probability density functions (pdf) for the NID variables.

The paper is organized as follows. In Section \ref{sec:NIDfull}, we propose our ``Latent Normalized Infinitely Divisible Topic Models'' and present its generative process. We dedicate Section \ref{sec:prop} to the properties of NID distributions and indicate how they overcome the drawbacks of the Dirichlet distribution and other distributions on the simplex. In Section \ref{sec:learning} we present our efficient learning algorithm with guaranteed convergence for the proposed topic model based on spectral decomposition. Finally, we conclude the paper in Section \ref{sec:conclusion}.

 \section{Latent Normalized Infinitely Divisible Topic Models}
\label{sec:NIDfull}

Topic models incorporate relationships between words $\bfx_1, \bfx_2\ldots\in \Rbb^d$  and a set of $k$ hidden topics. We represent the words $\bfx_i$ using one-hot encoding, i.e. $\bfx_i = \bfe_j$ if the $j^{\tha}$ word in the vocabulary occurs, and $\bfe_j$ is the standard basis vector.   The proportions of topics in a document is represented by vector $\hv \in \Rbb^k$. We assume that $\hv$ is drawn from an NID distribution. 

The detailed generative process of a latent NID topic model for each document is as follows
\begin{itemize}
	\item[1.] Draw $k$ independent variables, $z_1, z_2, \ldots, z_k$ from a family of ID distributions.
	\item[2.] Set $\hv$ to $ (\frac{z_1}{Z}, \dots, \frac{z_k}{Z})$ where $Z = \sum_{i \in [k]}z_i$.
	\item[3.] For each word $\bfx_i $,
	\begin{itemize}
		\item[(a)] Choose a topic $\zeta_i \sim \text{Multi}(\hv)$ and represent it with one-hot encoding.
		\item[(b)] Choose a word $\bfx_i$ vector as a standard basis vector with probability 
\begin{equation}
\label{eq:lda1}
	\mathbb{E} (\xv_i \vert \zeta_i) = \Am \zeta_i,
\end{equation}  conditioned on the drawn topic $\zeta_i$, and $\bfA \in \Rbb^{d \times k}$ is the topic-word matrix.
	\end{itemize}
\end{itemize}

From \eqref{eq:lda1}, we also have \begin{equation}
\label{eq:lda}
	\mathbb{E} (\xv_i \vert \hv) =
	\Ebb[\mathbb{E} (\xv_i \vert  \hv,\zeta_i) ]=
	\mathbb{E} (\xv_i \vert \zeta_i) \Ebb(\zeta_i \vert \hv) = \Am \hv.
\end{equation}

When the $z_i$ is drawn from the Gamma$(\alpha_i,1)$ distribution, we obtain the Dir$(\alphabf)$ distribution for the hidden vector $\hv = (h_1,\dots, h_k)$, and the LDA model through the above generative process.

Our goal is to recover the topic-word matrix $\Am$ given the document collection. In the following section we introduce the class of NID distribution and discuss  its properties.

\section{Properties of NID distributions}
\label{sec:prop}
\begin{figure}[t]
\centering
\includegraphics[width=1.5in]{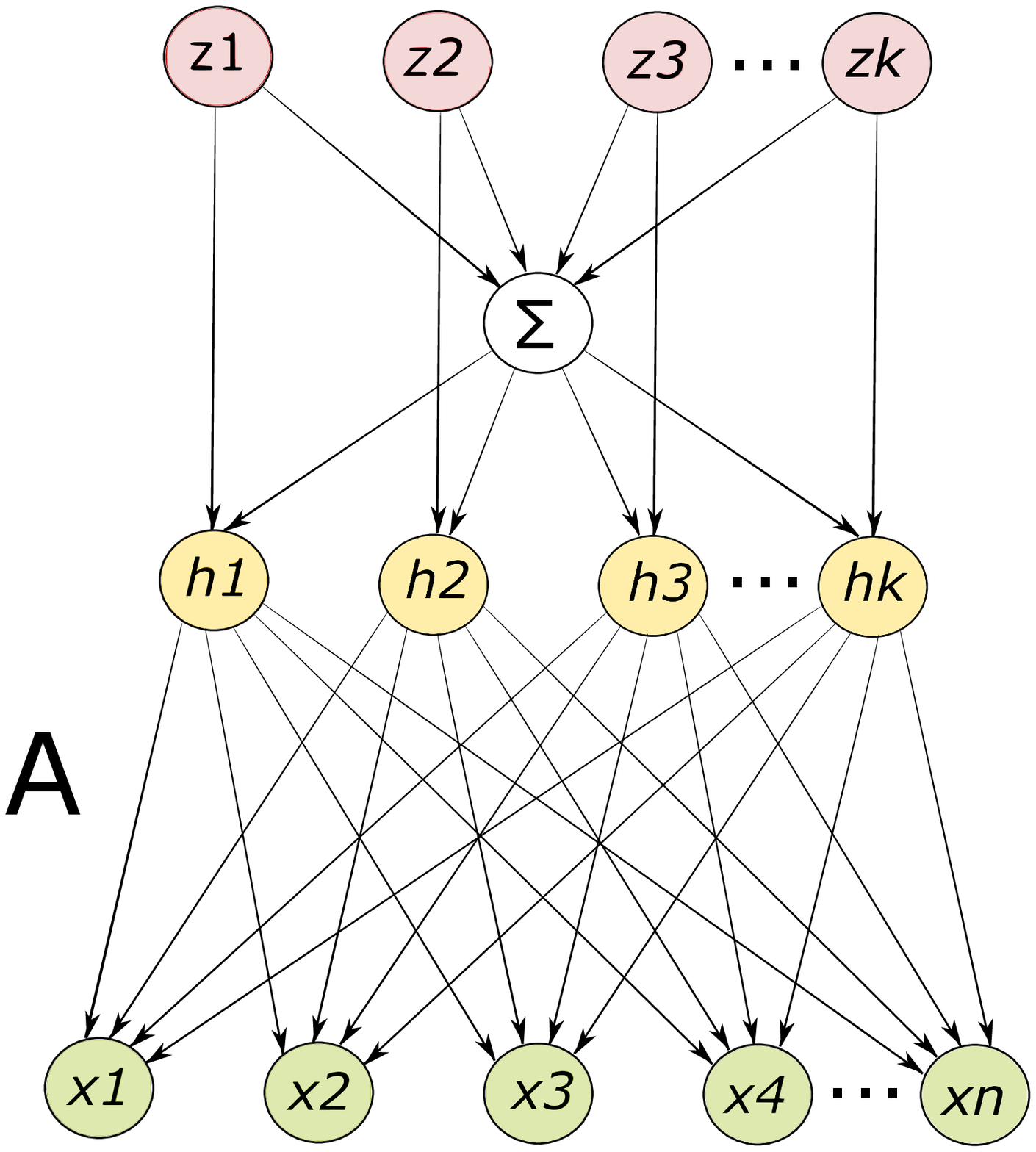}
\caption{Graphical Model Representation of the Latent NID Topic Model. $z_1,z_2,\ldots,z_k $ are a collection of independent Infinitely Divisible positive variables that are characterized by the collection of their corresponding L\'{e}vy measures $\alpha_1 \nu, \alpha_2 \nu, \dots, \alpha_k \nu$ And $h_1, h_2\ldots, h_k$ are the resulting NID variables representing topic proportions in a document of length $N$ with words $x_1, \dots, x_N$ }
\label{fig:graphical}
\end{figure}
 
NID distributions are a flexible class of distributions on the simplex and have been applied in a range of domains. This includes   hierarchical mixture modeling with Normalized Inverse-Gaussian distribution~\citep{lijoi2005hierarchical}, and modeling overdispersion with the normalized tempered stable distribution \cite{kolossiatis2011modeling}, both of which are examples of NID distributions.  For more applications, see \cite{favaro2011class}.
Let us first define the concept of infinite divisibility and present the properties of an ID distribution, and then consider the NID distributions.

\subsection{Infinitely Divisible Distributions}
\label{sec:ID}
If random variable $z$ has an Infinitely Divisible (ID) distribution, then for any $n \in \mathbb{N}$ there exists a collection of i.i.d random variables $y_1, \dots , y_n$ such that $z \stackrel{\text{d}}{=} y_1 + \dots + y_n$. In other words, an Infinitely Divisible distribution can be expressed as the sum of an arbitrary number of independent identically distributed random variables.

The Poisson distribution, compound Poisson, the negative binomial distribution,  Gamma distribution, and the trivially degenerate distribution are examples of Infinitely Divisible distributions; as are the normal distribution, Cauchy distribution, and all other members of the stable distribution family. The Student's t-distribution is also another example of Infinitely Divisible distributions. The uniform distribution and the binomial distribution are not infinitely divisible, as are all distributions with bounded (finite) support.

The special decomposition form of ID distributions makes them natural choices for certain models or applications. E.g. a compound Poisson distribution is a Poisson sum of IID random variables. The discrete compound Poisson distribution, also known as the stuttering Poisson distribution, can model batch arrivals (such as in a bulk queue~\citep{adelson1966compound}) and can incorporate Poisson mixtures.

In the sequel, we limit the discussion to ID distributions on $\mathbb{R}^+$ in order to ensure that the Normalized ID variables are on the simplex.  Let us now present how ID distributions can be characterized.

\paragraph{L\'{e}vy measure:} A $\sigma$-finite Borel measure $\nu$ on $\mathbb{R^{+}}$ is called a L\'{e}vy measure if $\int_0^\infty \text{min}(1,x) \nu(\mathrm{d} x) < \infty$. According to the L\'{e}vy-Khintchine representation given below, the L\'{e}vy measure uniquely characterizes an ID distribution along with a constant scale $\tau$. This implies  that every Infinitely Divisible distribution corresponds to a L\'{e}vy process, which is a stochastic process with independent increments.

\paragraph{L\'{e}vy-Khintchine representation}[Theorem 16.14 \cite{klenke2014infinitely}] Let $\mathcal{M}_1(\Lambda)$ and $\mathcal{M}_\sigma(\Lambda)$ indicate the set of probability measures and the set of $\sigma$-finite measures on a non-empty set $\Lambda$, respectively. Let $\mu\in \mathcal{M}_1([0,\infty))$ and let $\Psi(u) = -\log \int \limits_0^\infty e^{-u z} \mathrm{d} (\mu)$ be the log-Laplace transform of $\mu$. Then $\mu$ is Infinitely Divisible, if and only if there exists a $\tau \geq 0$ and a $\sigma$-finite measure $\nu \in \mathcal{M}_{\sigma}((0,\infty))$ with
\begin{equation}
	\int \limits_0^\infty \text{min}(1,z) \nu(\mathrm{d} z) < \infty,
\end{equation}
such that
\begin{equation}
	\Psi(u) = \tau u + \int \limits_0^\infty (1-e^{-uz}) \nu(\mathrm{d} z) \qquad \text{for} \quad u \geq 0,
\end{equation}
In this case the pair $(\tau, \nu)$ is unique, $\nu$ is called the L\'{e}vy measure of $\mu$ and $\tau$ is called the deterministic part. It can be shown that $\tau = \text{sup}\{ z \geq 0 : \mu([0,z)) = 0\}$.

In particular, let $\Phi_{z_i}(u) = \mathbb{E}[e^{\iota u z_i}] = \int \limits_0^\infty e^{\iota u z_i} f(z_i) \mathrm{d} z_i $ indicate the characteristic function of an Infinitely Divisible random variable $z_i$ with pdf $f(z_i)$ and corresponding pair $(\tau_i,\nu_i)$, where $\iota $ is the imaginary unit. Based on the L\'{e}vy-Khintchine representation it holds that $\Phi_{z_i}(\iota u) = \mathbb{E}[e^{-u z_i}] = e^{ - \Psi_i(u)}$ where $\Psi_i(u) = \tau_i u + \int \limits_0^\infty (1 - e^{-u z}) \nu_i(\mathrm{d} z)$ is typically referred to as the Laplace exponent of $z_i$. This implies that the Laplace exponent of an ID variable is also completely characterized by pair $(\tau_i,\nu_i)$. It holds for ID variables that if $\nu_i$ is a well-defined L\'{e}vy measure, so is $\alpha_i \nu_i$ for any $\alpha_i > 0$, which indicates that $\alpha_i \Psi_i(u)$ is also a well-defined Laplace exponent of an ID variable.


%


\subsection{Normalized Infinitely Divisible Distributions}
\label{sec:NID}
As defined in \cite{favaro2011class}, a Normalized Infinitely Divisible (NID) random variable is a random variable that is formed by normalizing independent draws of strictly positive (not necessarily coinciding) Infinitely Divisible distributions. More specifically, let $z_1, \dots, z_k$ be a set of independent strictly positive Infinitely Divisible random variables and $Z = z_1 + \dots + z_k$. An NID distribution is defined as the distribution of the random vector $\hv = (h_1, \dots, h_k ) := (\frac{z_1}{Z}, \dots, \frac{z_k}{Z})$ on the $(k-1)$-dimensional simplex, denoted as $\Delta^{k-1}$. The strict positivity assumption implies that $\hv$ is on the simplex~\cite{favaro2011class, mangili2015new}.

Let $[k]$ denote Natural numbers ${1,\dots,k}$. As stated by the L\'{e}vy-Khintchine theorem, a collection of ID positive variables $z_i$ for $i \in [k]$ is completely characterized by the collection of the corresponding L\'{e}vy measures $\nu_1, \dots, \nu_k$. It was shown in \cite{mangili2015new} that this also holds for the normalized variables $h_i$ for $i \in [k]$.

In this paper, we assume that the ID variables $z_1,\dots,z_k$ are drawn independently from ID distributions that are characterized with the corresponding collection of L\'{e}vy measures $\alpha_i \nu, \dots, \alpha_k \nu$, respectively. Which in turn translates respectively to variables with Laplace exponents $\alpha_1 \Psi(u), \dots, \alpha_k \Psi(u)$. Variables $\alpha_i$ will allow the distribution to vary in the interior of the simplex, providing the asymmetry needed to model latent models. The homogeneity assumption on the L\'{e}vy measure or the Laplace exponent provides the structure needed for guaranteed learning (Theorem \ref{thm:mainResult}). The overall graphical model representation is shown in Figure~\ref{fig:graphical}
\begin{figure*}
\label{fig:simplex}
\centering
	\begin{subfigure}[b]{0.3\textwidth}
		\includegraphics[width=\textwidth]{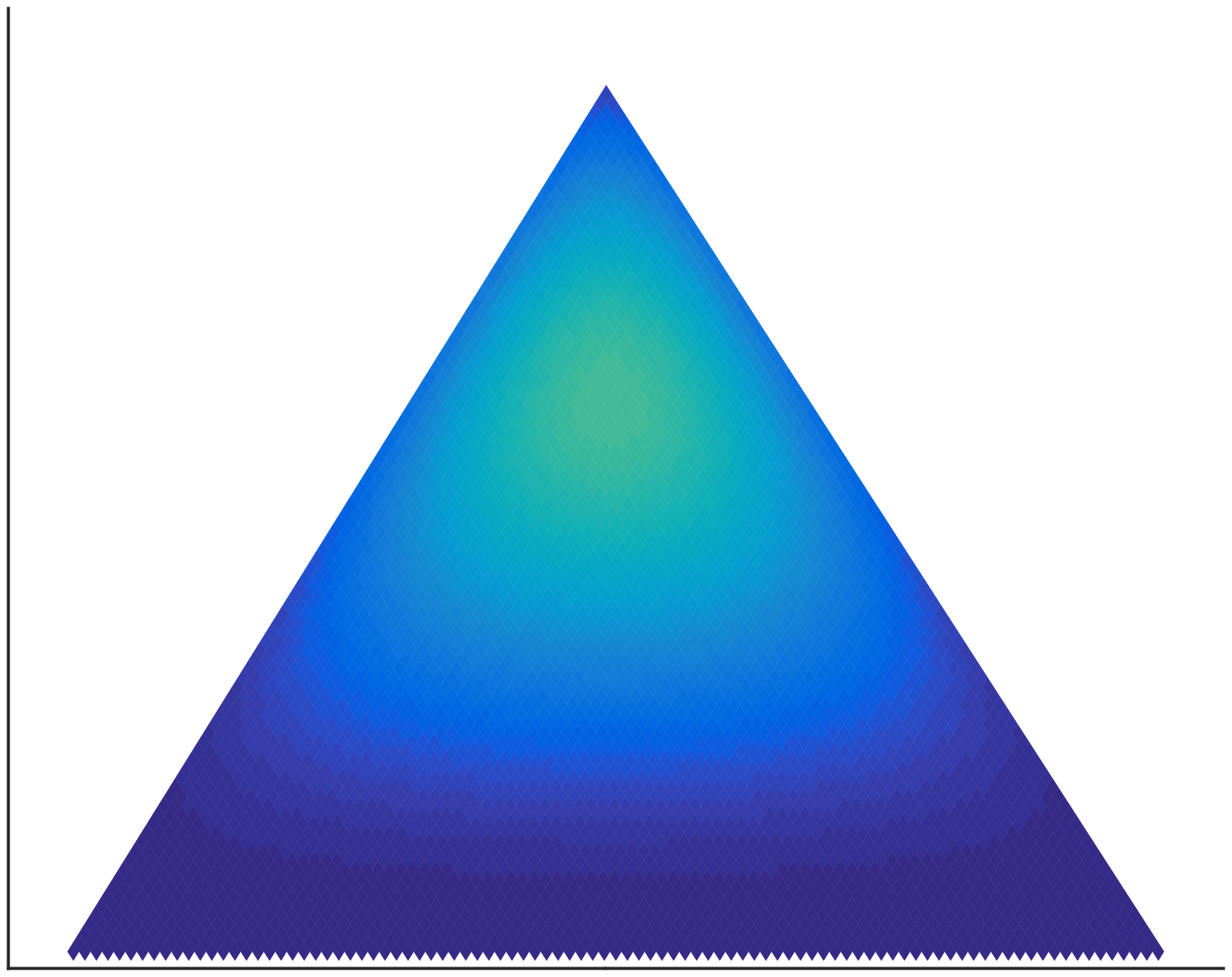}
		\caption{Gamma, $\lambda=1$}
		\label{fig:gamCent}
	\end{subfigure}
	\begin{subfigure}[b]{0.3\textwidth}
		\includegraphics[width=\textwidth]{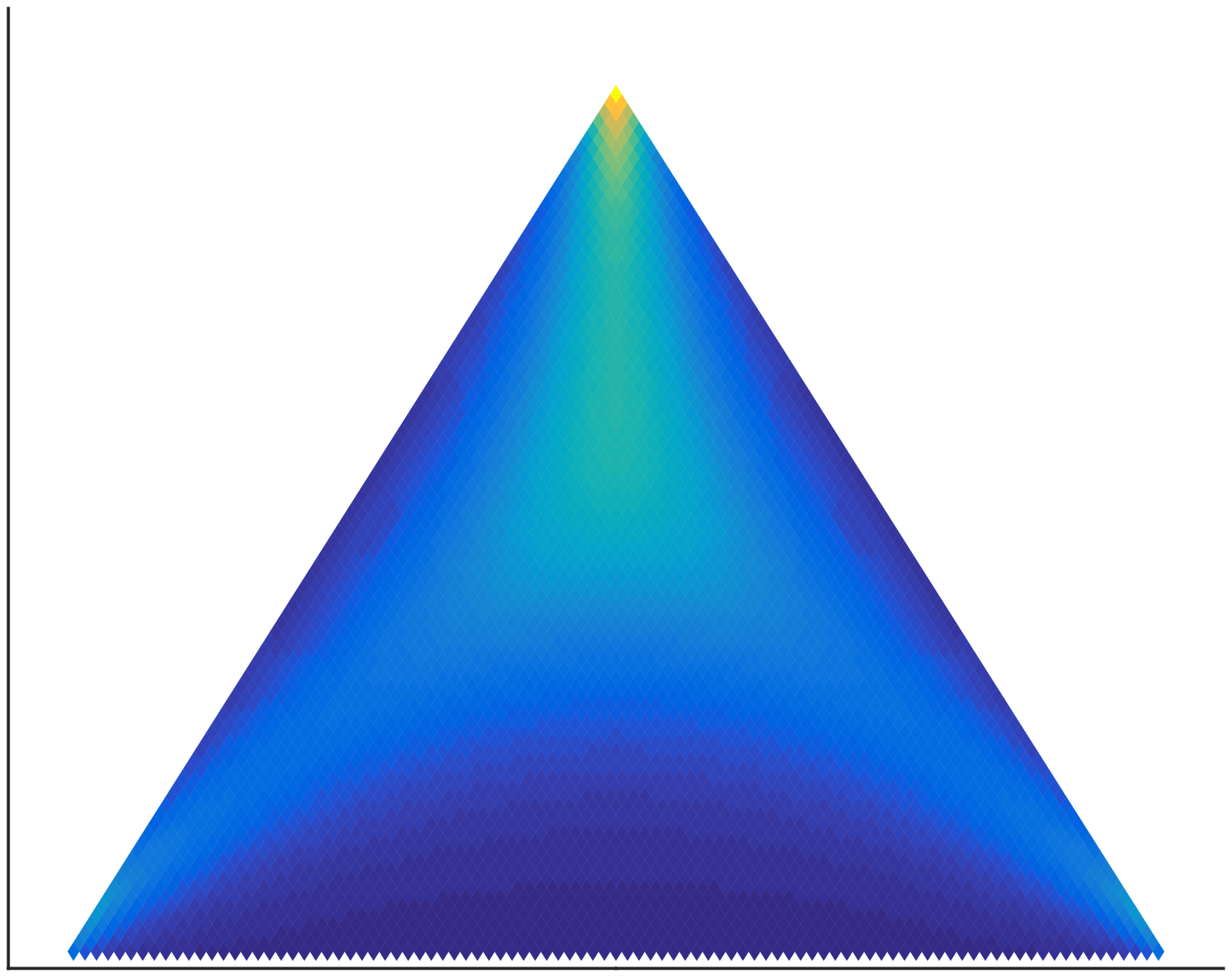}
		\caption{$\gamma$-stable, $\gamma=0.75$}
		\label{fig:stableCent}
	\end{subfigure}
	\begin{subfigure}[b]{0.32\textwidth}
		\includegraphics[width=\textwidth]{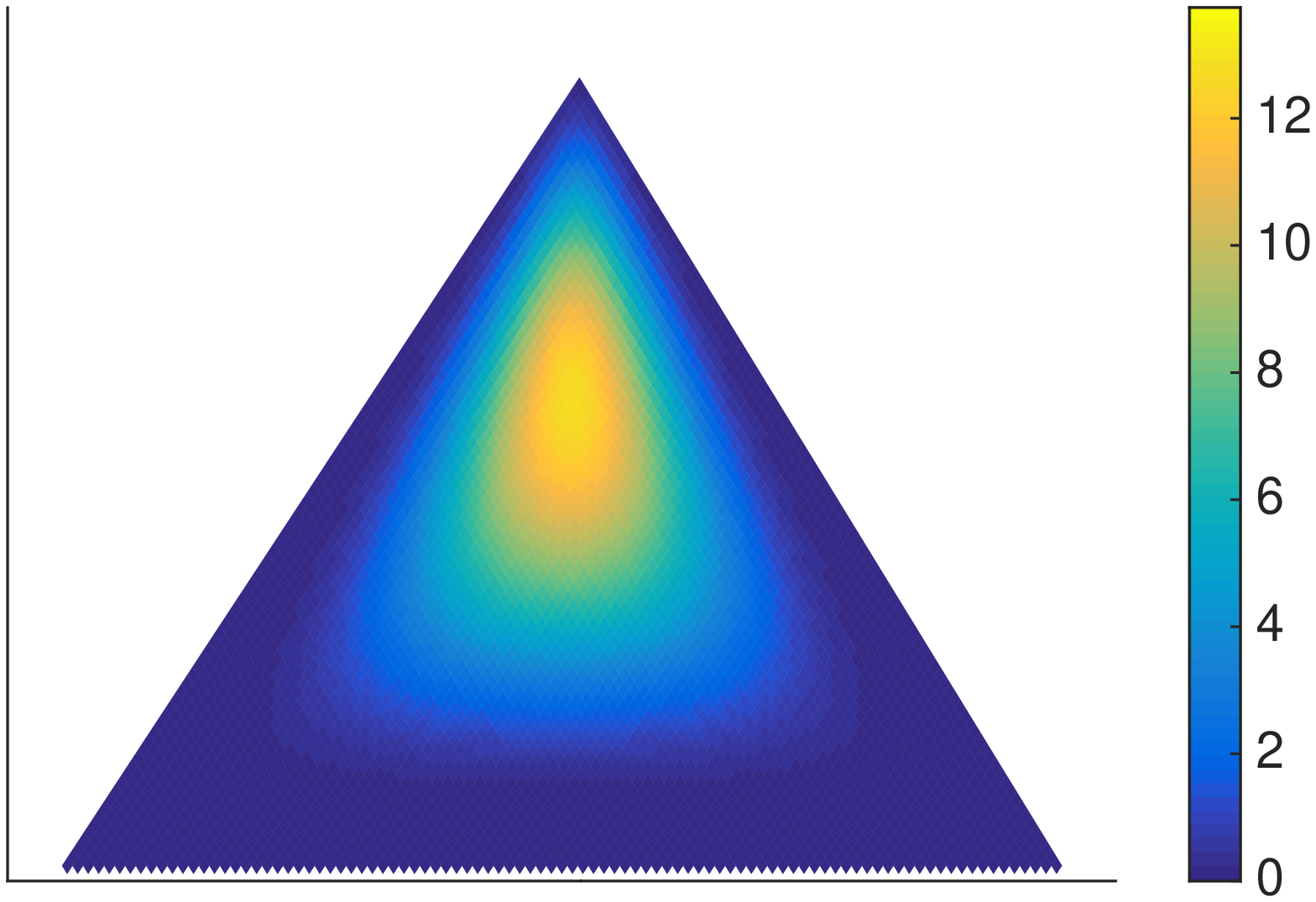}
		\caption{inverse Gaussian, $\lambda=0.01$}
		\label{fig:invGaussCent}
	\end{subfigure}
	\begin{subfigure}[b]{0.3\textwidth}
		\includegraphics[width=\textwidth]{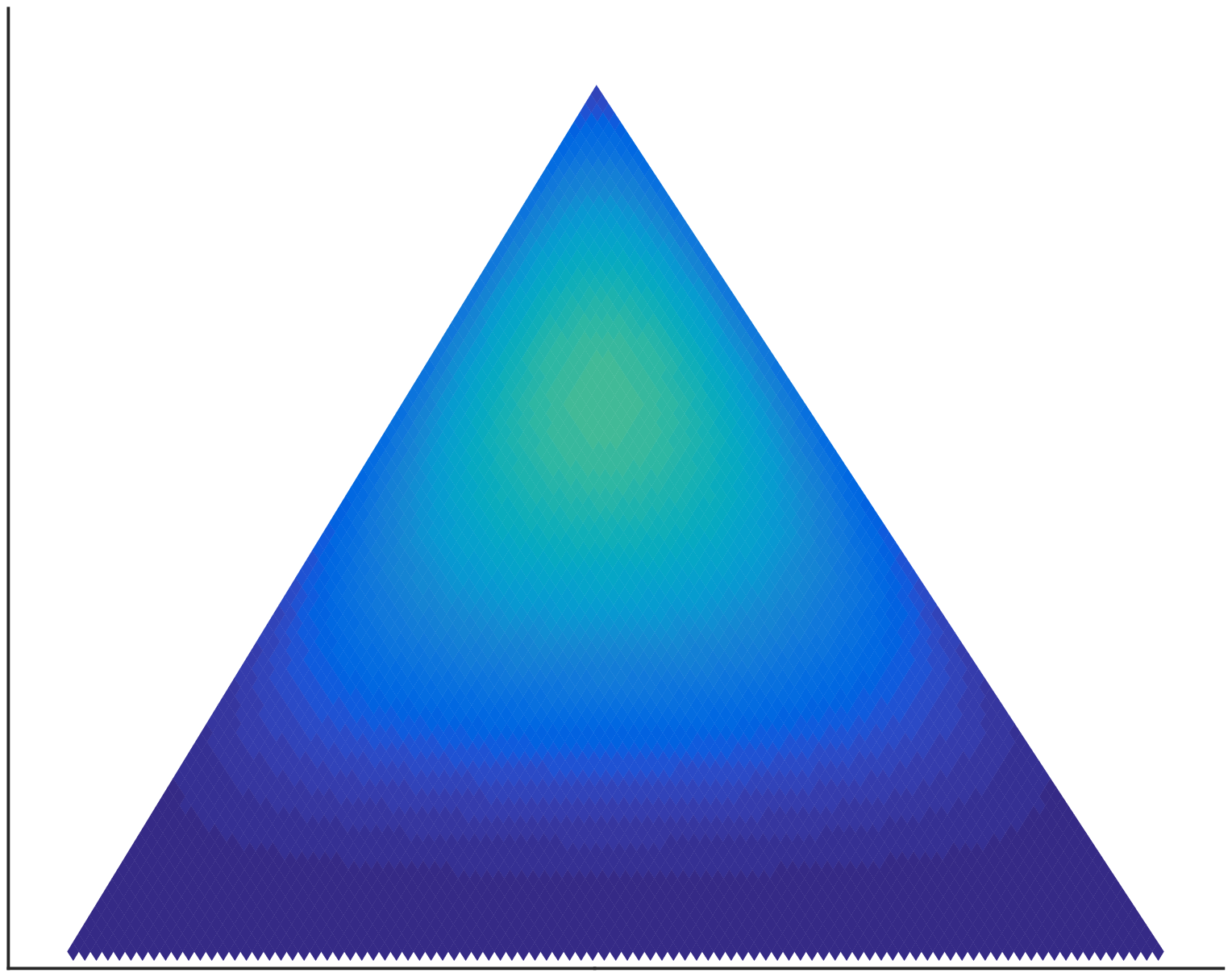}
		\caption{Gamma, $\lambda=10$}
		\label{fig:gamVert}
	\end{subfigure}
	\begin{subfigure}[b]{0.3\textwidth}
		\includegraphics[width=\textwidth]{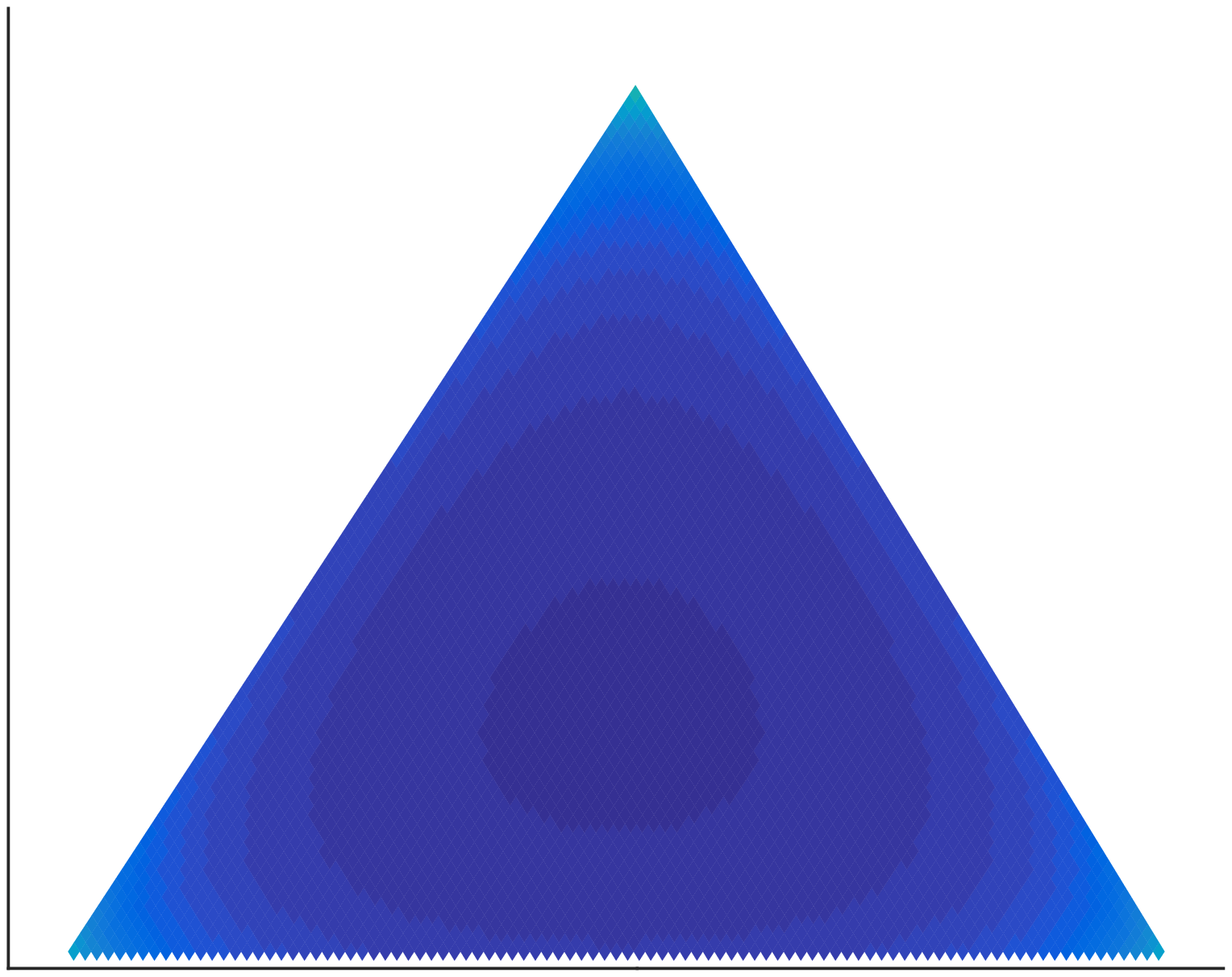}
		\caption{$\gamma$-stable, $\gamma=0.4$}
		\label{fig:stableVert}
	\end{subfigure}
	\begin{subfigure}[b]{0.32\textwidth}
		\includegraphics[width=\textwidth]{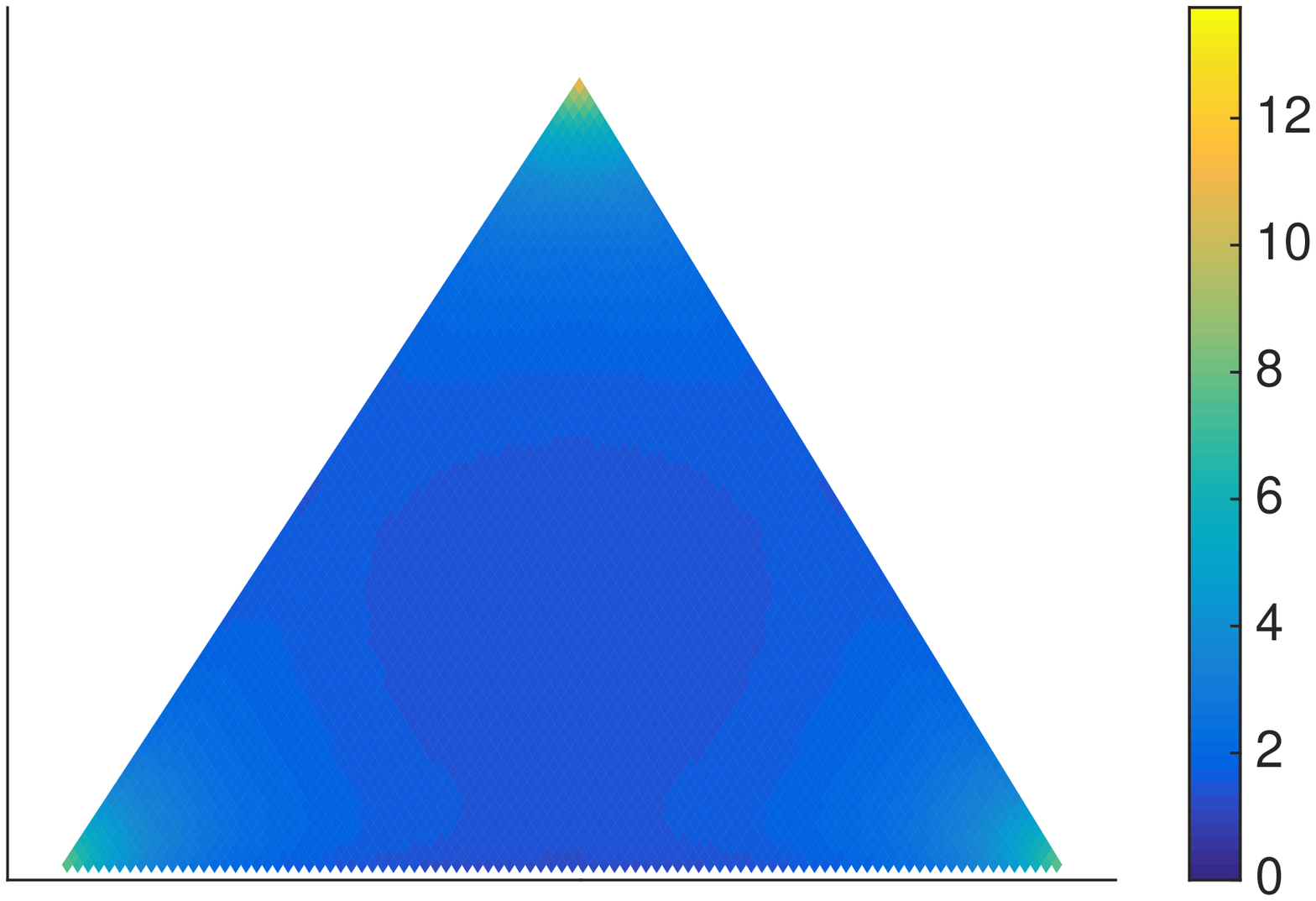}
		\caption{inverse Gaussian, $\lambda=4$}
		\label{fig:invGaussVert}
	\end{subfigure}
	\caption{Heat map of the pdf of three examples of the NID class that have closed form with respect to their parameters. All the figures have $\alpha = (2,2,4)$. 
	For the Inverse Gaussian the distribution moves from the center to the vertices of the simplex as $\lambda$ goes from $0$ to $\infty$ with fixed $\alphabf$ and for the $\gamma$-stable we have the same behavior when $\gamma$ changes from $1$ to $0$ with fixed $\alphabf$.}
	\label{fig:simplices}
\end{figure*}

If the original ID variables $z_i$ have probability densities $f_i$ for all $i \in [k]$, then the distribution of vector $\hv$, where $h_k = 1 - \sum_{i \in [k-1]} h_i$ is,
$
	f(\hv) = \int \limits_0^\infty \prod \limits_{i \in [k]} f_i(h_i Z) Z^{k-1} \mathrm{d} Z.
$
There are only three members of the NID class that have closed form densities namely, the Gamma distribution, Gamma$(\alpha_i, \lambda)$, the Inverse Gaussian distribution, $IG(\alpha_i,\lambda)$, and the $1/2$-stable distribution $St(\gamma,\beta,\alpha_i,\mu)$ with $\gamma=1/2$. $\mu = 1$ and $\beta = 1$ to ensure positive support for the Stable distribution. As noted earlier, Gamma$(\alpha_i, 1)$ reduces to the Dirichlet distribution. An interested reader is referred to \cite{favaro2011class,mangili2015new} for the closed form of each distribution.

Figure \ref{fig:simplices} depicts the heatmap of the density of these distributions on the probability simplex for different value of their parameters. Note that all the distributions have the same $\alpha$ parameter and hence, the same mean values. However, their concentration properties are widely varying, showing that the NID class can incorporate variations in higher order moments through additional parameters.

\paragraph{Gamma ID distribution: }
When the ID distribution is Gamma with parameters $(\alpha_i,1)$, we have the Dirichlet distribution as the resulting NID distribution. The Laplace exponent for this distribution will, therefore, be $\Psi_i(u) = \alpha_i \text{ln}(1+u).$
\paragraph{$\gamma$-stable ID distribution: }
The variables are drawn from the positive stable distribution $St(\gamma, \beta, \alpha_i, \mu)$ with $\mu = 0$, $\beta = 1$ and $\gamma <1$ which ensures that the distribution is on $\mathbb{R}^+$. The Laplace exponent of this distribution is $\Psi_i(u) = \alpha_i \frac{\Gamma(1-\gamma)}{\sqrt{2 \pi}\gamma}u^\gamma.$
Note that the $\gamma$-stable distribution can be represented in closed form for $\gamma = \frac{1}{2}$.

\paragraph{Inverse Gaussian ID distribution: }
The random variables are drawn from the Inverse-Gaussian (IG) distribution $IG(\alpha_i,\lambda)$. The Laplace exponent of this distribution is $\Psi_i(u) = \alpha_i \big( \sqrt{2 u + \lambda^2} - \lambda \big).$

\emph{Note:} The Dirichlet distribution, the $1/2$-Stable distribution and the Inverse Gaussian distribution are all special cases of the generalized Inverse Gaussian distribution \cite{favaro2011class}.

\begin{figure*}
\label{fig:corr}
\centering
	\begin{subfigure}[b]{0.32\textwidth}
		\psfrag{lambda}[l]{$\lambda$}
		\psfrag{Positively Correlated Proportion}[l]{}
		\includegraphics[width=\textwidth]{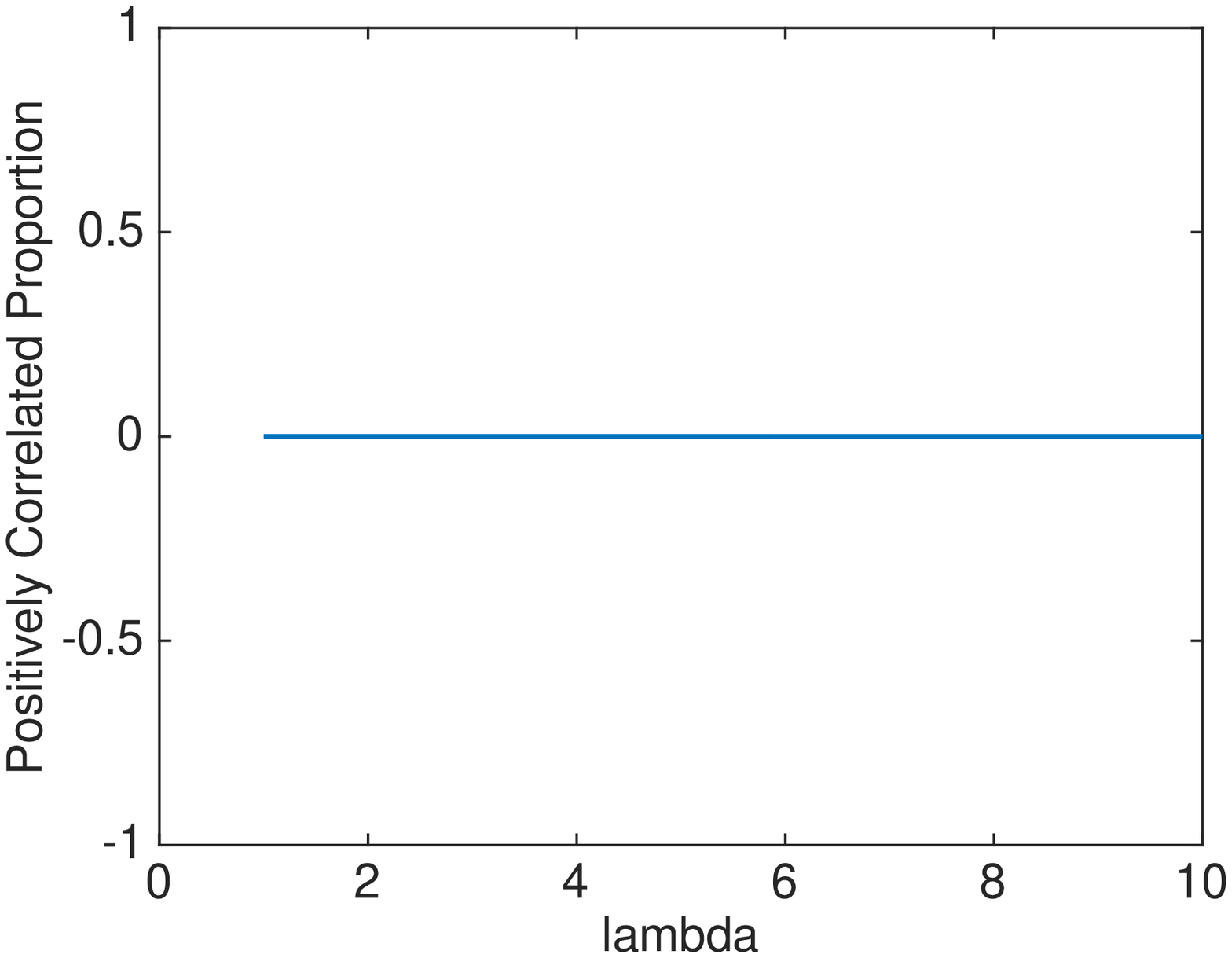}
		\caption{Gamma NID.}
		\label{fig:gamCorr}
	\end{subfigure}
	\begin{subfigure}[b]{0.32\textwidth}
		\psfrag{mu}[l]{$\lambda$}
		\psfrag{Positively Correlated Proportion}[l]{}
		\includegraphics[width=\textwidth]{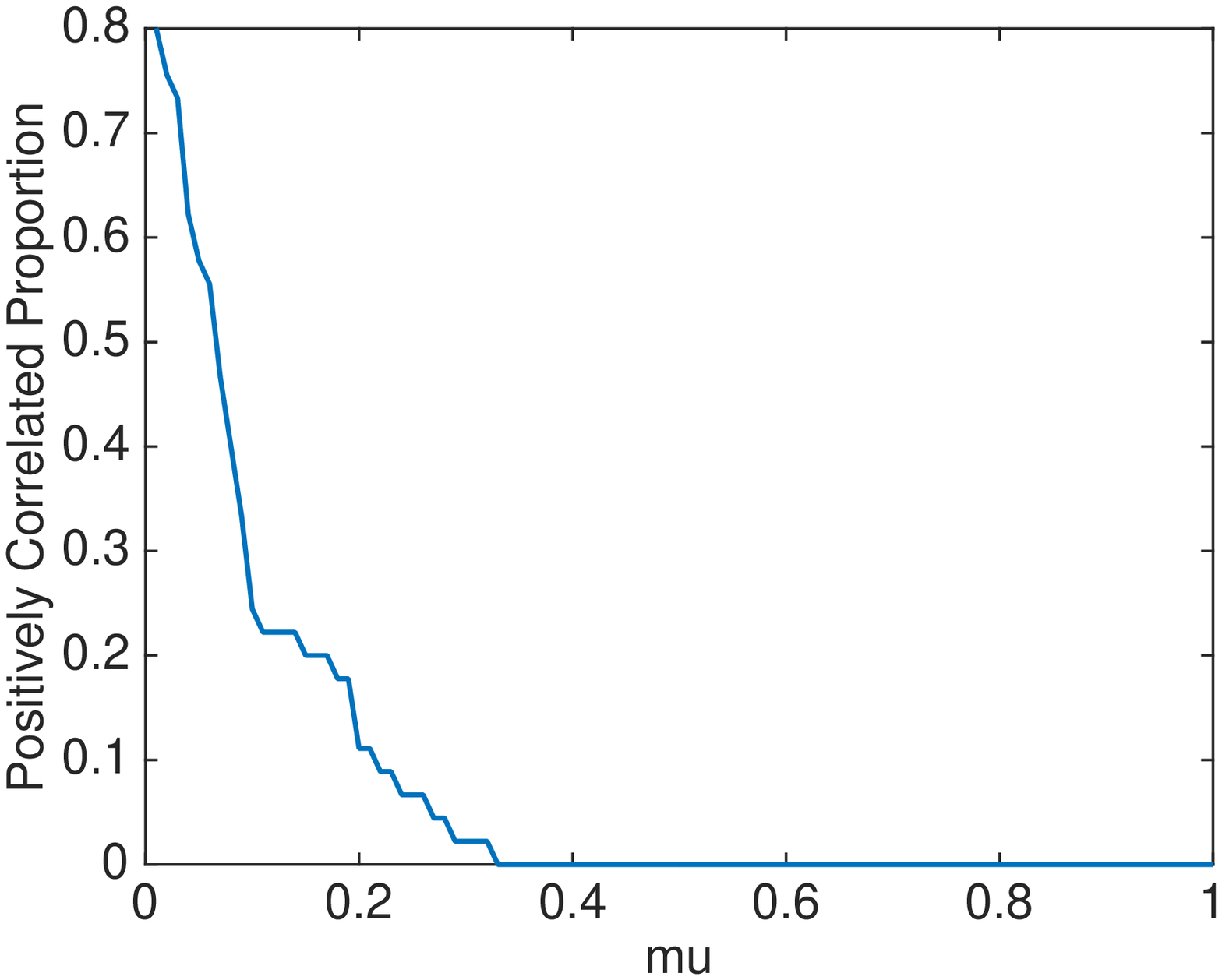}
		\caption{Inverse Gaussian NID.}
		\label{fig:invGaussCorr}
	\end{subfigure}
	\begin{subfigure}[b]{0.32\textwidth}
		\psfrag{alpha}[l]{$\gamma$}
		\psfrag{Positively Correlated Proportion}[l]{}
		\includegraphics[width=\textwidth]{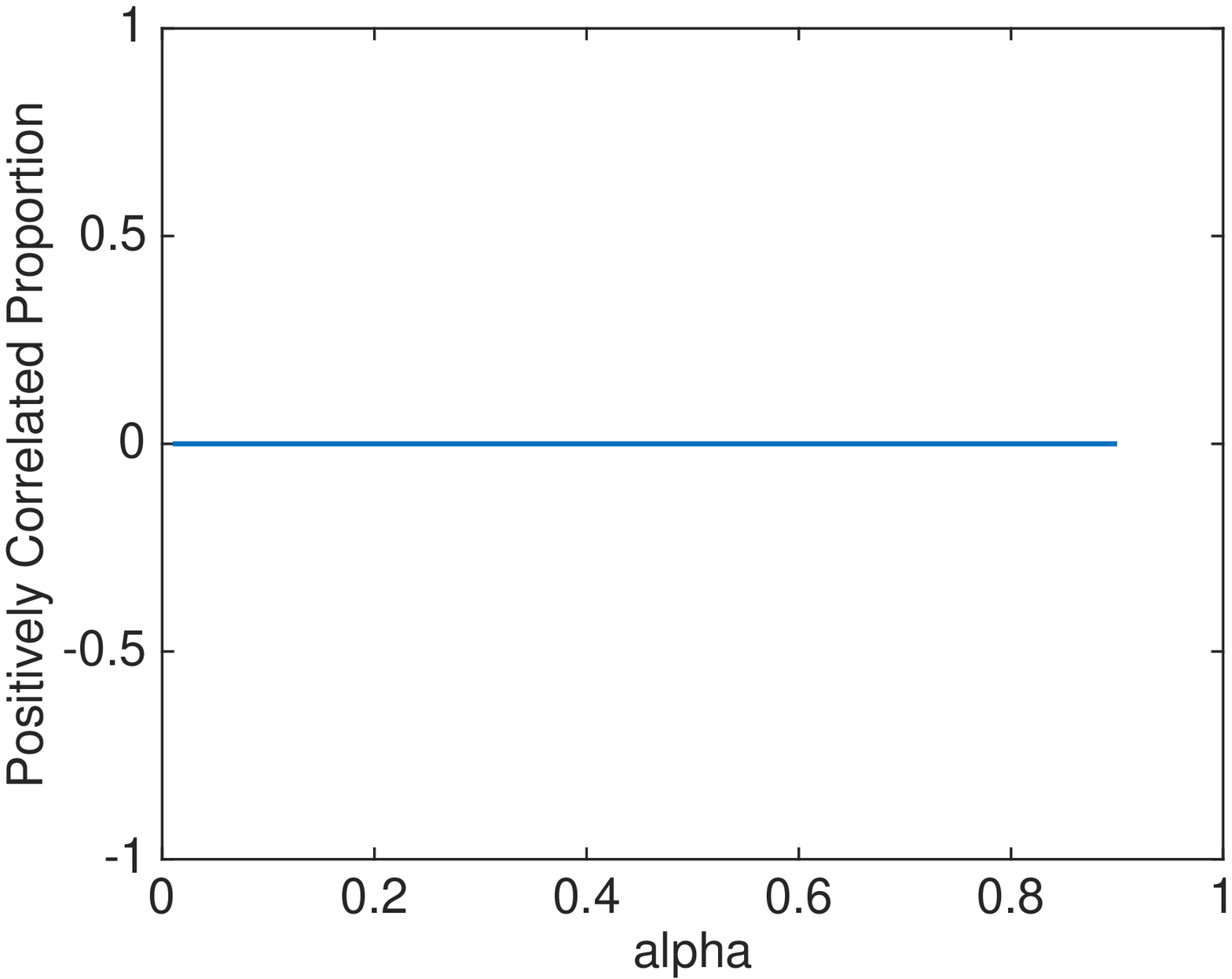}
		\caption{$\gamma$-stable NID.}
		\label{fig:gamStabCorr}
	\end{subfigure}
	\caption{Proportion of positively correlated elements of special cases of an NID distribution with 10 elements with respect to the parameter of the Laplace exponent for a fixed randomly drawn vector $\alphabf = [0.77,    0.70,    0.97, 0.46, 0.02, 0.44, 0.90,  0.33, 0.97, 0.45]$.}
	\label{fig:correlation}
\end{figure*}

As mentioned earlier, the class of NID distributions is capable of modeling positive and negative correlations among the topics. This property is depicted in Figure \ref{fig:correlation}. These figures show the proportion of positively correlated topics for the three presented distributions. As we can see the Inverse Gaussian NID distribution can capture both positive and negative correlations.

\section{Learning NID Topic Models through Spectral Methods}
\label{sec:learning}
In this section we will show how the form of the moments of NID distributions enable efficient learning of this flexible  class. 


In order to be able to guarantee efficient learning using higher order moments, the moments need to have a very specific structure. Namely, the moment of the underlying distribution of $\hv$ needs to form a diagonal tensor. If the components of $\hv$ where indeed independent, this is obtained through the cumulant tensor. On the other hand, for LDA, it has been shown by Anandkumar et. al. \cite{anandkumar2012spectral} that a linear combination of moments of up to third order of $\hv$ forms a diagonal tensor for the Dirichlet distribution. Below, we extend the result to the more general class of NID distributions.

\subsection{Consistency of Learning through Moment Matching}
\begin{assumption}
\label{as:homog}
	ID random variables $z_i$ for $i \in [k]$ are said to be  \emph{partially homogeneous} if they   share the same L\'{e}vy measure. This implies   that the corresponding Laplace exponent of variable $z_i$ is given $ \alpha_i \Psi(u)$ for some $\alpha_i \in \Rbb^+$, and $\Psi(u)$ is the Laplace exponent of the common L\'{e}vy measure.
\end{assumption}

Under the above assumption, we prove guaranteed learning of NID models through spectral methods. This is based on the following moment forms for NID models, which admit a CP tensor decomposition. The components of the decomposition will be the columns of the topic-word matrix: $\bfA:=[\bfa_1| \bfa_2|  \ldots |\bfa_k]$.

Define \beq
	\Omega  (m,n,p) = \int \limits_0^\infty u^m \frac{\mathrm{d}^n}{\mathrm d u^n} \Psi(u) \Big( \frac{\mathrm{d}}{\mathrm d u} \Psi(u) \Big)^p e^{- \alpha_0 \Psi(u)} \mathrm{d} u,\eeq where $ \Psi(u)$ is the Laplace exponent of the NID distribution and $\alpha_0 = \sum_{i \in [k]} \alpha_i$.

\begin{theorem}({\textbf{Moment Forms for NID models}})
\label{thm:mainResult}
Let ${\mathbf{M}_2}$ and ${\mathbf{M}_3}$ be respectively the following matrix and tensor constructed from the following  moments of the data,
\begin{align}
	{\mathbf{M}_2} = & \mathbb{E}[\xv_1 \otimes \xv_2] +  v \cdot \mathbb{E}[\xv_1] \otimes \mathbb{E}[\xv_2] ,  \label{eq:momentP} \\
	{\mathbf{M}_3} = & \mathbb{E}[\xv_1 \otimes \xv_2 \otimes \xv_3] + v_2 \cdot \mathbb{E}[\xv_1] \otimes \mathbb{E}[\xv_2] \otimes \mathbb{E}[\xv_3]   \nonumber \\
	&+ v_1 \cdot \big[\mathbb{E}[\xv_1 \otimes \xv_2] \otimes \mathbb{E}[\xv_3] + \nonumber \\
	 & \qquad \quad \mathbb{E}[\xv_1] \otimes \mathbb{E}[\xv_2 \otimes \xv_3] +  \nonumber \\
	  & \qquad \quad \mathbb{E}[\xv_1 \otimes \mathbb{E}[\xv_2] \otimes \xv_3 ]  \big] \label{eq:momentT} \\
\end{align}
where,
\begin{align}
	v &= \frac{\Omega(1,1,1)}{\Big( \Omega(0,1,0) \Big)^2}, \quad
	v_1 = - \frac{\Omega(2,2,1)}{2 \Omega(1,2,0) \Omega(0,1,0)}, \quad  \label{eq:v}  \\
	v_2& = \frac{-0.5 \Omega(2,1,2) + 3 v_1 \Omega(1,1,1) \Omega(0,1,0)}{\Big( \Omega(0,1,0) \Big)^3}, \label{eq:vv} 
\end{align}
 Then given Assumption \ref{as:homog},
\begin{align}
	{\mathbf{M}_2} = \sum \limits_{j \in [k]} \kappa_j (\mathbf{a}_j \otimes \mathbf{a}_j),  \quad 
	{\mathbf{M}_3} = \sum \limits_{j \in [k]} \lambda_j (\mathbf{a}_j \otimes \mathbf{a}_j \otimes \mathbf{a}_j). \label{eq:3rdDecomp}
\end{align}
for a set of $ \kappa_j$'s and $\lambda_j$'s which are a function of the parameters of the distribution.
\end{theorem}

\paragraph{Remark 1: efficient computation of $v, v_1$ and $v_2$: }What makes Theorem \ref{thm:mainResult} specially intriguing is the fact that weights $v$, $v_1$ and $v_2$ can be computed through univariate integration, which can be computed efficiently, regardless of the dimensionality of the problem.  



\paragraph{Remark 2: investigation of special cases}
When the ID distribution is Gamma with parameters $(\alpha_i,1)$, we have the Dirichlet distribution as the resulting NID distribution.
Weights $v_1$ and $v_2$ reduce to the results of Anandkumar et. al. \cite{anandkumar2012spectral} for the Gamma$(\alpha_i,1)$ distribution,
which are $v_1  = - \frac{\alpha_0}{\alpha_0 + 2}$ and $v_2  = \frac{2 \alpha_0^2}{(\alpha_0+2)(\alpha_0+1)}.$
When the variables are drawn from the positive stable distribution $St(1/2, \beta, \alpha_i, \mu)$
weights $v_1$ and $v_2$ in Theorem \ref{thm:mainResult} can be represented in closed form as $v_1 = - \frac{1}{4}$ and $v_2 = -\frac{5}{8}$.

\begin{figure}
\centering
	\begin{subfigure}[b]{0.45\textwidth}
		\psfrag{v1}[l]{$v_1$}
		\psfrag{gamma}[l]{$\gamma$}
		\includegraphics[width=\textwidth]{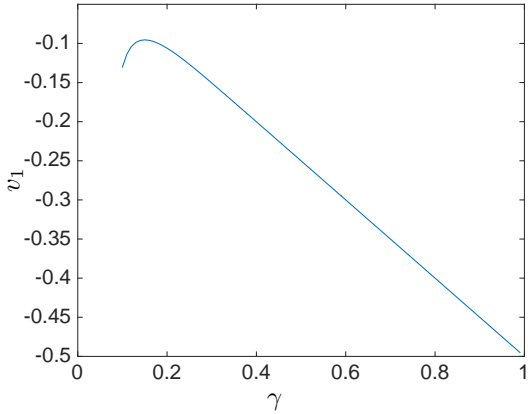}
		\caption{Weight $v_1$ of theorem \ref{thm:mainResult} for a Stable ID distribution $St(\gamma, \beta, \alpha_i, \mu)$ with $\mu=0$, $\beta=1$, $\gamma < 1$ and $\alpha_i > 0$ vs. $\gamma$ for $\alpha_0 = 1$}
		\label{fig:gamStab}
	\end{subfigure}
	\qquad
	\begin{subfigure}[b]{0.45\textwidth}
		\psfrag{v1}[l]{$v_1$}
		\psfrag{gamma}[l]{$\lambda$}
		\includegraphics[width=\textwidth]{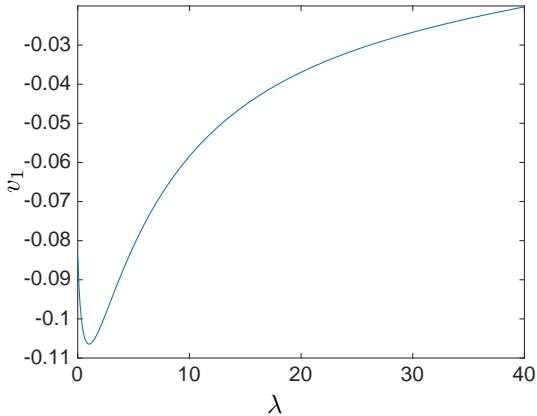}
		\caption{Weight $v_1$ of theorem \ref{thm:mainResult} for an Inverse Gaussian distribution $IG(\alpha_i,\lambda)$ vs. $\lambda > 0$ and $\alpha_i \geq 0$ for $\alpha_0 = 1$}
		\label{fig:invGauss}
	\end{subfigure}
	\caption{Weight $v_1$ for two different examples of the NID distribution. Weights $v$ and $v_2$ in the theorem have similar behavior w.r.p the parameters.}
	\label{fig:weights}
\end{figure}

It is hard to find closed form representation of the weights for other stable distributions and the Inverse Gaussian distribution. Therefore, we give the form of the weights with respect to the parameters of each distribution in Figure \ref{fig:weights}. As it can be seen in Figures \ref{fig:stableVert} and \ref{fig:stableCent}, as $\gamma$ increases, the distribution gets more centralized on the simplex. Therefore, as depicted in Figure \ref{fig:gamStab} the weight becomes more negative to compensate for it. The same holds in Figure \ref{fig:invGauss}.



The above result immediately implies guaranteed learning for non-degenerate topic-word matrix $\bfA$. 

\begin{assumption}
\label{as:lin}
Topic-word matrix $\bfA\in \Rbb^{d\times k}$ has linearly independent columns and the parameters $\alpha_i> 0$. 
\end{assumption}

\begin{corollary}
\textbf{\emph{(Guaranteed Learning of NID Topic Models using Spectral Methods)}}
Given empirical versions of moments $\bfM_2$ and $\bfM_3$ in \eqref{eq:momentP} and \eqref{eq:momentT}, using tensor decomposition algorithm from~\cite{anandkumar2014tensor}, under the above assumption, we can consistently estimate topic-word matrix $\Am$ and parameters $\alphabf$ with polynomial computational and sample complexity.
\end{corollary}
The overall procedure is given in Algorithm \ref{alg:learning}.

\paragraph{Remark 3: third order moments suffice} For the flexible class of latent NID topic models, only moments up to the third order suffice for efficient learning.

\begin{algorithm}
\caption{Parameter Learning}
\label{alg:learning}
\begin{algorithmic}[1]
\Require Chosen NID distribution and hidden dimension $k$
\Ensure Parameters of NID distribution $\alphabf$ and topic-word matrix $\Am$
\State Estimate empirical moments $\hat{\mathbb{E}}(\xv_1 \otimes \xv_2 \otimes \xv_3)$ ,$\hat{\mathbb{E}}(\xv_1 \otimes \xv_2)$ and $\hat{\mathbb{E}}(\xv_1)$.
\State Compute weights $v$, $v_1$ and $v_2$ in \eqref{eq:v} and \eqref{eq:vv} for the given NID distribution by numerical integration.
\State Estimate tensors $\mathbf{M}_2$ and $\mathbf{M}_3$ in \eqref{eq:momentP} and \eqref{eq:momentT} .
\State Decompose tensor $\mathbf{M}_3$ into its rank-$1$ components using the algorithm in \cite{anandkumar2014tensor} that requires $\mathbf{M}_2$.
\State Return columns of $\Am$  as the components of the decomposition.
\end{algorithmic}
\end{algorithm} 

\paragraph{Remark 4: Sample Complexity} Following \cite{anandkumar2012spectral}, Algorithm \ref{alg:learning} can recover matrix $\Am$ under Assumption \ref{as:lin} with polynomial sample complexity.

\paragraph{Remark 5: Implementation Efficiency} In order to make the implementation efficient we use the discussion in \cite{anandkumar2014tensor}. Specifically, as mentioned in \cite{anandkumar2014tensor}, we can find a whitening transformation from matrix ${\mathbf{M}_2}$ that lowers the data dimension from the vocabulary space to the topic space. We then use the same whitening transformation to go back to the original space and recover the parameters of the model.

\paragraph{Overview of the proof of Theorem~\ref{thm:mainResult}}

We begin the proof by forming the following second order and third order tensors using the moments of the NID distribution given in Lemma \ref{lem:moment}.
\begin{align}
	{\mathbf{M}_2^{(\hv)}} & = \mathbb{E}(\hv \otimes \hv) + v \mathbb{E}(\hv) \otimes \mathbb{E}(\hv) ,\label{eq:2ndMom} \\
	{\mathbf{M}_3^{(\hv)}} & = \mathbb{E}(\hv \otimes \hv \otimes \hv) + v_2  \mathbb{E}(\hv) \otimes \mathbb{E}(\hv) \otimes \mathbb{E}(\hv) \nonumber \\
	&+ v_1 \mathbb{E}(\hv \otimes \hv) \otimes \mathbb{E}(\hv) \nonumber \\
	& + v_1 \mathbb{E}(\hv \otimes \mathbb{E}(\hv) \otimes \hv) \nonumber \\
	& + v_1 \mathbb{E}(\hv) \otimes \mathbb{E}(\hv \otimes \hv) \label{eq:3rdMom} 
\end{align}
Weights $v$, $v_1$ and $v_2$ are as in Equations \eqref{eq:v} and \eqref{eq:vv}. They are  computed by setting the off-diagonal entries of matrix ${\mathbf{M}_2^{(\hv)}}$ in Equation \ref{eq:2ndMom} and ${\mathbf{M}_3^{(\hv)}}$ in Equation \ref{eq:3rdMom} to $0$.  Due to the homogeneity assumption, all the off-diagonal entries can be simultaneously made to vanish with these choices of coefficients for $v, v_1$ and $v_2$. We obtain 
 ${\mathbf{M}_2^{(\hv)}} = \sum_{i \in [k]} \kappa'_i \mathbf{e}_i^{\otimes 2} $ and ${\mathbf{M}_3^{(\hv)}} = \sum_{i \in [k]} \lambda'_i \mathbf{e}_i^{\otimes 3}$ where $\mathbf{e}_i$'s are the standard basis vectors, and this implies they are diagonal tensors. Due to this fact and the exchangeability of the words given topics according to \eqref{eq:lda}, Equations \ref{eq:3rdDecomp} follow.

%
The exact forms of $v, v_1$ and $v_2$ are obtained by the following moment forms for NID distributions.

\begin{lemma}[\cite{mangili2015new}]
\label{lem:moment} The moments of NID variables $h_1, \ldots h_k$ satisfy
\begin{equation}
\label{eq:NIDmoments}
		\mathbb{E}(h_1^{r_1} h_2^{r_2} \dots h_k^{r_k}) = \frac{1}{\Gamma(r)} \int \limits_0^\infty u^{r-1} e^{-\alpha_0 \Psi(u)} \prod \limits_{j \in [k]} B_{r_j}^j \mathrm{d} u,
\end{equation}
where $r = \sum_{i \in  [k]} r_i$ and $B_{r_j}^j$ can be written in terms of the partial Bell polynomial as
\begin{equation}
	B_{r}^i = B_r(-\alpha_i\Psi^{(1)}(u), \dots, -\alpha_i\Psi^{(r)}(u)),
\end{equation}
in which $\Psi^{(l)}(u)$ is the $l$-th derivative of $\Psi(u)$ with respect to $u$.
\end{lemma}



\section{Experiments}
In this section we apply our proposed latent NID topic modeling algorithm to New York Times and Pubmed articles \cite{Lichman:2013}. The New York Times dataset contains about $300,000$ documents and the pubmed data contains around $8$ million documents. The vocabulary size for both the datasets are around $100,000$. 

\begin{centering}
\begin{table}
\centering
\caption{Top 10 Words for Pubmed, K = 10}
\label{tab:topWordsPubmed}
\scalebox{0.8}{
	\begin{tabularx}{0.5\textwidth}{| c || X |}
	\hline
	Topic & Top Words in descending order of importance \\
 	\hline
1 & protein, region, dna, family, sequence, gene, form-12, analysis.abstract, model, tumoural \\ 
2 & cell, mice.abstract, expression.abstract, activity.abstract, primary, tumor, antigen, human, t-cell, vitro \\ 
3 & tumor, treatment, receptor, lesional, children--a, effect.abstract, factor, rat1, renal-cell, response-1 \\ 
4 & patient, treatment, therapy, clinical, disease, level.abstract, effect.abstract, treated, tumor, surgery \\ 
5 & activity.abstract, rat1, concentration, dna, human, effect.abstract, exposure.abstract, animal-based, reactional, inhibition.abstract \\ 
6 & patient, children--a, women.abstract, treatment, level.abstract, syndrome, disordered, disease, year-1, therapy \\ 
7 & effect.abstract, receptor, level.abstract, rat1, mutational, gene, concentration, women.abstract, insulin, expression.abstract \\ 
8 & acid, strain, concentration, women.abstract, test, pregnancy--a, drug, system--a, function.abstract, water \\ 
9 & strain, protein, system--a, muscle, mutational, species, growth, diagnosis-based, analysis.abstract, gene \\ 
10 & infection.abstract, hospital, programed, strain, medical, alpha, information, health, children--a, data.abstract \\ 
 	\hline
	\end{tabularx}
	}
\end{table}
\end{centering}

\begin{centering}
\begin{table*}
\caption{NID Top 10 Words for NYtimes, K = 20}
\label{tab:topWords}
\scalebox{1}{
	\begin{tabularx}{\textwidth}{| c || X |}
	\hline
	Topic & Top Words in descending order of importance \\
 	\hline
 	1 & seeded, soldier, firestone, bobby-braswell, michigan-state, actresses, gary-william, preview, school-district, netanyahu \\ 
2 & diane, question, newspaper, copy, fall, held, tonight, send, guard, slugged \\ 
3 & abides, acclimate, acetate, alderman, analogues, annexing, ansar, antitax, antitobacco, argyle \\ 
4 & percent, school, quarter, company, taliban, high, stock, race, companies, john-mccain \\ 
5 & test, deal, contract, tiger-wood, question, houston-chronicle, copy, won, seattle-post-intelligencer ,tax \\ 
6 & tonight, diane, question, newspaper, file, copy, fall, slugged, onlytest, xxx \\ 
7 & company, com, market, stock, won, los-angeles-daily-new, business, eastern, web, commentary \\ 
8 & abides, acclimate, acetate, alderman, analogues, annexing, ansar, antitax, antitobacco, argyle \\ 
9 & company, game, run, los-angeles-daily-new, percent, team, season, stock, companies, games \\ 
10 & working-girl, abides, acclimate, acetate, alderman, analogues, annexing, ansar, antitax, antitobacco \\ 
11 & diane, newspaper, fall, tonight, question, held, copy, bush, slugged, police \\ 
12 & hurricanes, policies, surgery, productivity, courageous, emergency, singapore, orange-bowl, regarding, telecast \\ 
13 & abides, acclimate, acetate, alderman, analogues, annexing, ansar, antitax, antitobacco, argyle \\ 
14 & company, com, won, stock, market, eastern, commentary, business, web, deal \\ 
15 & company, stock, market, business, investor, technology, analyst, cash, sell, executives \\ 
16 & tonight, question, diane, file, newspaper, copy, fall, slugged, onlytest, xxx \\ 
17 & defense, held, children, fight, assistant, surgery, michael-bloomberg, worker, bird, omar \\ 
18 & percent, company, stock, companies, quarter, school, market, analyst, high, corp \\ 
19 & school, student, yard, released, guard, premature, teacher, touchdown, publication, leader \\ 
20 & school, percent, student, yard, high, taliban, flight, air, afghanistan, plan \\ 
 	\hline
	\end{tabularx}
	}
\end{table*}
\end{centering}

\begin{centering}
\begin{table}[ht]
\centering
\caption{Perplexity comparison accross different datasets}
\label{tab:perplexity}
	\begin{tabular}{|c||c|c|}
 	\hline
	Dataset & NYtimes & Pubmed \\
	\hline
	NID & $\mathbf{3.5702e+03}$ &  $\mathbf{4.0771e+03}$ \\
	LDA & $4.8464e+03$ & $4.3702e+03$  \\
 	\hline
	\end{tabular}
\end{table}
\end{centering}

\begin{centering}
\begin{table}[ht]
\centering
\caption{PMI comparison accross different datasets}
\label{tab:PMI}
	\begin{tabular}{|c||c|c|}
 	\hline
	Dataset & NYtimes & Pubmed \\
	\hline
	NID & $\mathbf{0.2439}$ &  ${0.3080}$ \\
	LDA & $0.2362$ & $\mathbf{0.4487}$  \\
 	\hline
	\end{tabular}
\end{table}
\end{centering}

\begin{centering}
\begin{table}[ht]
\centering
\caption{10 Shared words: New York times dataset}
\label{tab:sharedWords}
\scalebox{0.8}{
	\begin{tabularx}{0.60\textwidth}{| c || X |}
 	\hline
 	Shared Words & boston-globe, tonight, question, newspaper, spot, percent, file, diane, copy, fall \\
 	 	\hline
	\end{tabularx}
	}
\end{table}
\end{centering}

\paragraph{Hyperparameter Tuning} In practice, we can tune for hyperparameters to compute the best fitting $v, v_1$ and $v_2$. Therefore, we will not limit ourselves to a single parametric NID family. We learn the weights during the learning process and employ a non-parametric estimation of the L\'{e}vy-Khintchine representation through the univariate integrals of Equations \ref{eq:v} and \ref{eq:vv}. Due to the one-dimensional nature of the integrations, a small number of parameters will suffice for good performance. The following paragraph describes the process in more detail.

We first split the data into train and test sets randomly. We then use the train data to learn the model parameters, $\alpha_i$'s and the columns of the topic-word matrix $\Am$, as well as the weights $v$, $v_1$ and $v_2$ in Equations \ref{eq:momentP} and \ref{eq:momentT}, respectively. We do so by finding the best low rank approximation of tensor ${\mathbf{M}_3} $ that minimizes the Frobenius Norm difference between the right-hand-side of Equation \ref{eq:momentT} and its low rank approximation. The recovered components are the columns of the topic-word matrix and the parameters $\alpha_i$ are recovered from the decomposition weights. Once we find the best $v$, $v_1$ and $v_2$ we use the test data to find the best NID distribution described by the weights such that the likelihood of the test data is maximized under that choice of NID distribution. 

\paragraph{Results:}
We compare our proposed latent NID topic model with the spectral LDA method \cite{anandkumar2012spectral}. It has been shown in \cite{huang2016discovery} that spectral LDA is more efficient and achieves better perplexity compared to the conventional LDA \cite{blei2003latent}. Table \ref{tab:topWords} provides a sketch of the top words per topics recovered by our latent NID topic model on the New Yowk times dataset and Table \ref{tab:topWordsPubmed} shows the top words recovered from the pubmed dataset. We have also provided the the top words recovered by LDA for the New York times dataset for comparison purposes in Table \ref{tab:topWordslda} in the appendix. Besides from the top words, we also present the shared words among the recovered topics for the New York Times dataset in Table \ref{tab:sharedWords}. The presence of words such as ``tonight'', ``question'' and ``fall'' among these words makes a lot of sense since they are general words that are not usually indicative of any specific topic.

\begin{figure}[ht]
\centering
	\begin{subfigure}[b]{0.45\textwidth}
		\includegraphics[width=\textwidth]{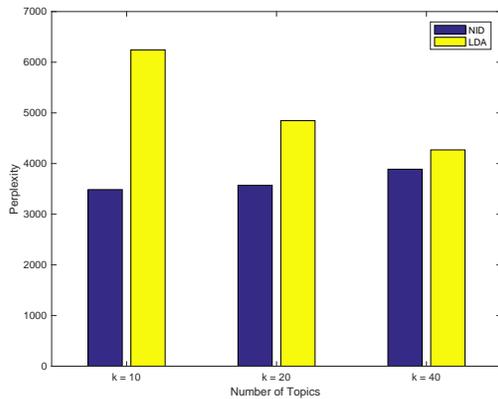}
		\caption{Perplexity score}
		\label{fig:gamCent}
	\end{subfigure}
	\begin{subfigure}[b]{0.45\textwidth}
		\includegraphics[width=\textwidth]{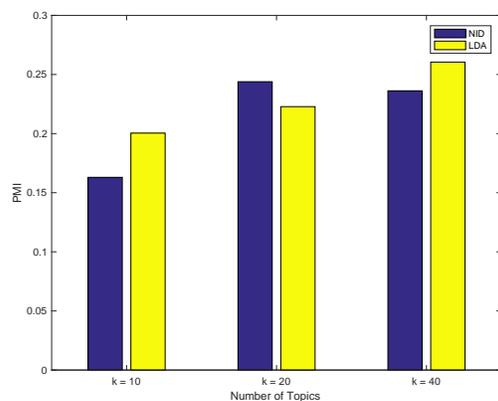}
		\caption{Pointwise Mutual Information (PMI)}
		\label{fig:stableCent}
	\end{subfigure}
	\caption{Perplexity and PMI score for the NYtimes dataset across different number of topics}
	\label{fig:perp}
\end{figure}

We use the well-known likelihood perplexity measure \cite{blei2003latent} to evaluate the generalization performance of our proposed topic modeling algorithm as well as the Pointwise Mutual Information (PMI) score \cite{anandkumar2013learning} to assess the coherence of the recovered topics. Perplexity is defined as the inverse of the geometric mean per-word of the estimated likelihood. We refer to our proposed method as $NID$ and compare it against $LDA$ \cite{anandkumar2012spectral} where the distribution of the hidden space is fixed to be Dirichlet. It should be noted that lower perplexity indicates better generalization performance and higher PMI indicated better topic coherence. Figure \ref{fig:perp} shows the perplexity and PMI score for the NID and LDA methods across different number of topics for the New York Times dataset. Similar comparisons including the Pubmed dataset results are also provided in Tables \ref{tab:perplexity} and \ref{tab:PMI}. The results suggest that if we allow the corpus to choose the best underlying topic distribution, we can get better generalization performance as well as topic coherence on the held-out set compared to fixing the underlying distribution to Dirichlet. The improved perplexity of our proposed method is indicative of correlations in the underlying documents that are not captured by the Dirichlet distribution. Thus, latent NID topic models are capable of successfully capturing correlations within topics while providing guarantees for exact recovery and efficient learning as proven in Section \ref{sec:learning}. 

Last but not least, the naive Variational Inference implementation of \cite{blei2003latent} \footnote{available at: http://www.cs.princeton.edu/~blei/lda-c/}, does not scale to the current datasets used in this paper. The naive implementation of the spectral LDA, however, takes only about a minute to run on the NYtimes dataset and about 15 minutes to run on the Pubmed dataset. It is, therefore, of great importance to have a class of models that can be learned using spectral methods mainly because of their inherent scalability, ease of implementation and statistical guarantees. As we show in this paper, latent NID topic models are such a class of models. The correlated topic model framework of \cite{blei2006correlated} also uses Variational Inference to perform learning and it is limited to the logit-normal distribution. latent NID topic models are not only scalable, but are also capable of modeling arbitrary correlations without requiring a fixed prior distribution on the topic space.

\section{Conclusion}\label{sec:conclusion}
In this paper we introduce the new class of Latent Normalized Infinitely Divisible (NID) topic models that generalize previously proposed topic models such as LDA. We provide guaranteed efficient learning for this class of distributions using spectral methods through untangling the dependence of the   hidden topics. We provide evidence that our proposed NID topic model overcomes the shortcomings of the Dirichlet distribution by allowing for both positive and negative correlations among the topics. 
In the end we use two real world datasets to validate our claims in practice.
The improved likelihood perplexity score indicates that if we allow the model to pick the underlying distribution we will get better generalization results.


\clearpage
\section*{Appendix}

\begin{centering}
\begin{table*}
\caption{LDA Top 10 Words for NYtimes, K = 20}
\label{tab:topWordslda}
\scalebox{1}{
	\begin{tabularx}{\textwidth}{| c || X |}
	\hline
	Topic & Top Words in descending order of importance \\
 	\hline
 	1&newspaper, question, copy, fall, diane, chante-lagon, kill, mandatory, drug, patient \\ 
2&held, guard, send, publication, released, advisory, premature, attn-editor, undatelined, washington-datelined \\ 
3&los-angeles-daily-new, slugged, com, xxx, www, x-x-x, web, information, site, eastern \\ 
4&million, shares, offering, boston-globe, debt, public, initial, player, bill, contract \\ 
5&onlytest, point, tax, case, court, lawyer, police, minutes, death, shot \\ 
6&held, released, publication, guard, advisory, premature, send, attn-editor, undatelined, washington-datelined \\ 
7&com, information, www, web, eastern, daily, commentary, business, separate, marked \\ 
8&boston-globe, spot, file, killed, tonight, women, earlier, article, george-bush, incorrectly \\ 
9&million, shares, offering, debt, public, initial, player, contract, bond, revenue \\ 
10 & boston-globe, spot, file, held, killed, attn-editor, earlier, article, court, women \\ 
11 & percent, market, stock, point, quarter, economy, rate, women, growth, companies \\ 
12 & boston-globe, spot, file, tonight, killed, earlier, article, women, incorrectly, news-feature \\ 
13 & held, guard, publication, released, send, advisory, premature, attn-editor, undatelined, washington-datelined \\ 
14 & los-angeles-daily-new, slugged, xxx, new-york, x-x-x, fund, bush, goal, king, evening \\ 
15 & tonight, copy, question, diane, fall, newspaper, russia, terrorist, russian, black \\ 
16 & slugged, los-angeles-daily-new, xxx, new-york, x-x-x, bush, run, school, inning, student \\ 
17 & onlytest, file, film, onlyendpar, movie, new-york, seattle-pi, los-angeles, sport, patient \\ 
18 & los-angeles-daily-new, slugged, xxx, x-x-x, student, inning, send, program, enron, game \\ 
19 & los-angeles-daily-new, slugged, xxx, new-york, x-x-x, fund, evening, program, student, enron \\ 
20 & test, houston-chronicle, hearst-news-service, seattle-post-intelligencer, ignore, patient, kansas-city, yard, race, doctor \\ 

 	\hline
	\end{tabularx}
	}
\end{table*}
\end{centering}

\paragraph{Proof of Theorem \ref{thm:mainResult}}
\begin{myproof}
	The moment form of Lemma \ref{lem:moment} can be represented as \cite{mangili2015new},
\begin{align}
	& \mathbb{E}(h_1^{r_1} h_2^{r_2} \dots h_n^{r_n}) = \nonumber \\
	&  \frac{1}{\Gamma(r)} \int \limits_0^\infty u^{r-1} e^{-\sum \limits_{i=n+1}^k \Psi_i(u)} \prod \limits_{j \in [n]} (-1)^{r_j} \frac{\mathrm{d}^{r_j}}{\mathrm{d} u^{r_j}} e^{- \Psi_j(u)} \mathrm{d} u. \label{eq:momApp}
\end{align}
We use the above general form of the moments to compute and diagonalize the following moment tensors,
\begin{align}
	{\mathbf{M}_2^{(\hv)}} & = \mathbb{E}(\hv \otimes \hv) + \eta \mathbb{E}(\hv) \otimes \mathbb{E}(\hv) , \label{eq:2ndMomApp}\\
	{\mathbf{M}_3^{(\hv)}} & = \mathbb{E}(\hv \otimes \hv \otimes \hv) 
	 \nonumber \\
	&+ \eta_1 \mathbb{E}(\hv \otimes \hv) \otimes \mathbb{E}(\hv) \nonumber \\
	&+  \eta_2 \mathbb{E}(\hv \otimes \mathbb{E}(\hv) \otimes \hv) \nonumber \\
	&+ \eta_3 \mathbb{E}(\hv) \otimes \mathbb{E}(\hv \otimes \hv) \nonumber \\
	&+ \eta_4  \mathbb{E}(\hv) \otimes \mathbb{E}(\hv) \otimes \mathbb{E}(\hv) .\label{eq:3rdMomApp}
\end{align}
Setting the off-diagonal entries of Equations \eqref{eq:2ndMomApp} and \eqref{eq:3rdMomApp} to $0$ and get the following set of equations
\begin{align}
	\mathbb{E}&(h_i h_j) + \eta \mathbb{E}(h_i) \mathbb{E}(h_j) = 0 \qquad \text{for } \quad i \neq j , \\
 	\mathbb{E}&(h_i h_j h_l) \nonumber \\
 	& + \eta_1 \mathbb{E}(h_i h_j) \mathbb{E}(h_l) \nonumber \\
 	& + \eta_2 \mathbb{E}(h_i h_l) \mathbb{E}(h_j) \nonumber \\
 	& + \eta_3 \mathbb{E}(h_j h_l) \mathbb{E}(h_i) \nonumber \\
 	& +  \eta_4 \mathbb{E}(h_i) \mathbb{E}(h_j) \mathbb{E}(h_l) =0 \nonumber \\
 	&  \qquad \qquad \qquad   \text{for} \quad i \neq j \neq l = 0,  \\
 	\mathbb{E}& (h^2_i h_l) \nonumber \\
 	& +  \eta_1 \mathbb{E}(h^2_i) \mathbb{E}(h_l) \nonumber \\
 	& + \eta_2 \mathbb{E}(h_i h_l) \mathbb{E}(h_i) \nonumber \\
 	& + \eta_3 \mathbb{E}(h_i h_l) \mathbb{E}(h_i) \nonumber \\
 	& + \eta_4 \mathbb{E}(h_i) \mathbb{E}(h_i) \mathbb{E}(h_l) = 0 \nonumber \\ 
 	& \qquad \qquad \qquad  \text{for} \quad i \neq l .
\end{align}
Writing the moments using Equation \eqref{eq:momApp}, assuming $\Phi_i(u) = \alpha_i \Psi(u)$, we get the following weights by some simple algebraic manipulations,
\begin{align}
	\eta & =\frac{\int \limits_0^\infty u e^{-\alpha_0 \Psi(u)} \big( \frac{\mathrm{d}}{\mathrm d u} \Psi(u) \big)^2 \mathrm{d} u}{\Big( \int \limits_0^\infty e^{-\alpha_0 \Psi(u)}  \frac{\mathrm{d}}{\mathrm d u} \Psi(u) \mathrm{d} u \Big)^2 }  \label{eq:eta} \\
	\eta_1 & = \eta_2 =\eta_3 \nonumber \\
	&= - \frac{\frac{1}{2} \int \limits_0^\infty u^2 e^{- \alpha_0 \Psi(u)} \frac{\mathrm{d}^2}{\mathrm d u^2} \Psi(u) \frac{\mathrm{d}}{\mathrm d u} \Psi(u) \mathrm{d} u}{ \int \limits_0^\infty u e^{- \alpha_0 \Psi(u)} \frac{\mathrm{d}^2}{\mathrm d u^2} \Psi(u) \mathrm{d} u \int \limits_0^\infty e^{- \alpha_0 \Psi(u)}  \frac{\mathrm{d}}{\mathrm d u} \Psi(u) \mathrm{d} u} \label{eq:eta1} \\
	\eta_4 &= \frac{f(\psi(u))}{\Big( \int \limits_0^\infty e^{- \alpha_0 \Psi(u)}  \frac{\mathrm{d}}{\mathrm d u} \Psi(u) \mathrm{d} u \Big)^3} \label{eq:eta2}
\end{align}
Where
\begin{align}
	 f(\psi(u))  & = -\frac{1}{2} \int \limits_0^\infty u^2 e^{- \alpha_0 \Psi(u)} \big(\frac{\mathrm{d}}{\mathrm d u} \Psi(u) \big)^3 \mathrm{d} u \nonumber \\
	 & + (\eta_1 + \eta_2 + \eta_3) \int \limits_0^\infty u e^{- \alpha_0 \Psi(u)} \big(\frac{\mathrm{d}}{\mathrm d u} \Psi(u) \big)^2 \mathrm{d} u \nonumber \\
	 & \cdot \int \limits_0^\infty e^{- \alpha_0 \Psi(u)}  \frac{\mathrm{d}}{\mathrm d u} \Psi(u) \mathrm{d} u
\end{align}
Setting $v = \eta$, $v_1 = \eta_1 = \eta_2 =\eta_3$ and $v_2 = \eta_4$ and defining 
\begin{equation}
	\Omega (m,n,p) := \int \limits_0^\infty u^m \frac{\mathrm{d}^n}{\mathrm d u^n} \Psi(u) \Big( \frac{\mathrm{d}}{\mathrm d u} \Psi(u) \Big)^p e^{- \alpha_0 \Psi(u)} \mathrm{d} u,
\end{equation}
the set of weights $v$, $v_1$ and $v_2$ have the following form,
\begin{align}
	v  &= \frac{\Omega(1,1,1)}{\Big( \Omega(0,1,0) \Big)^2}, \label{eq:v12} \\
	v_1& = - \frac{\Omega(2,2,1)}{2 \Omega(1,2,0) \Omega(0,1,0)}, \label{eq:v1} \\
	v_2 &= \frac{-0.5 \Omega(2,1,2) + 3 v_1 \Omega(1,1,1) \Omega(0,1,0)}{\Big( \Omega(0,1,0) \Big)^3}. \label{eq:v2} \\
\end{align}
Weights $v$, $v_1$ and $v_2$ ensure that moment tensors ${\mathbf{M}_2^{(\hv)}}$ and $	{\mathbf{M}_3^{(\hv)}}$ form diagonal tensors. Therefore they can be represented as,
\begin{align}
	{\mathbf{M}_2^{(\hv)}} & =  \sum_{i \in [k]} \kappa_i \mathbf{e}_i^{\otimes 2} , \\
	{\mathbf{M}_3^{(\hv)}} & = \sum_{i \in [k]} \lambda_i \mathbf{e}_i^{\otimes 3},
\end{align}
where,
\begin{align}
	\kappa_i &= \Ebb[ h_i^2 ] + v \Ebb[ h_i ]^2, \\
	\lambda_i & = \Ebb[ h_i^3 ]   + 3 v_1 \big( \Ebb[h_i^2] \Ebb[h_i] \big) + v_2 \big( \Ebb[ h_i ]^3 \big).
\end{align}

The exchangeability assumption on the word space gives,
\begin{equation}
\label{eq:mom1}
	 \mathbb{E} [ \xv_1 ] = \mathbb{E} \big( \mathbb{E} [ \xv_1  \vert \hv ] \big) = {\bf A} \mathbb{ E}(\hv),
\end{equation}
\begin{align}
\label{eq:mom2}
	 \mathbb{E} [ \xv_1 \otimes \xv_2]   = \mathbb{E} \big( \mathbb{E} [ \xv_1 \otimes \xv_2 \vert \hv ] \big) 
	  = {\bf A} \mathbb{E}(\hv \otimes \hv) {\bf A}^\top,
\end{align}
\begin{align}
\label{eq:mom3}
	 \mathbb{E} [ \xv_1 \otimes \xv_2 \otimes \xv_3 ] &  = \mathbb{E} \big( \mathbb{E} [ \xv_1 \otimes \xv_2 \otimes \xv_3 \vert \hv ] \big)\nonumber \\
	 &  = \mathbb{E} [ \hv \otimes \hv \otimes \hv] ({\bf A},{\bf A},{\bf A}).
\end{align}
Therefore,
\begin{align}
	{\mathbf{M}_2} &= \Am \mathbf{M}_2^{(\hv)} \Am^\top  = \sum \limits_{j \in [k]} \kappa_j (\mathbf{a}_j \otimes \mathbf{a}_j), \\
	{\mathbf{M}_3} &= \mathbf{M}_3^{(\hv)} (\Am, \Am, \Am)  = \sum \limits_{j \in [k]} \lambda_j (\mathbf{a}_j \otimes \mathbf{a}_j \otimes \mathbf{a}_j)
\end{align}
\end{myproof}


\begin{thebibliography}{10}

\bibitem{adelson1966compound}
RM~Adelson.
\newblock Compound poisson distributions.
\newblock {\em OR}, 17(1):73--75, 1966.

\bibitem{anandkumar2012spectral}
Anima Anandkumar, Yi-kai Liu, Daniel~J Hsu, Dean~P Foster, and Sham~M Kakade.
\newblock A spectral algorithm for latent dirichlet allocation.
\newblock In {\em Advances in Neural Information Processing Systems}, pages
  917--925, 2012.

\bibitem{anandkumar2014tensor}
Animashree Anandkumar, Rong Ge, Daniel Hsu, Sham~M Kakade, and Matus Telgarsky.
\newblock Tensor decompositions for learning latent variable models.
\newblock {\em The Journal of Machine Learning Research}, 15(1):2773--2832,
  2014.

\bibitem{anandkumar2013learning}
Animashree Anandkumar, Ragupathyraj Valluvan, et~al.
\newblock Learning loopy graphical models with latent variables: Efficient
  methods and guarantees.
\newblock {\em The Annals of Statistics}, 41(2):401--435, 2013.

\bibitem{arora2013practical}
Sanjeev Arora, Rong Ge, Yonatan Halpern, David~M Mimno, Ankur Moitra, David
  Sontag, Yichen Wu, and Michael Zhu.
\newblock A practical algorithm for topic modeling with provable guarantees.
\newblock In {\em ICML (2)}, pages 280--288, 2013.

\bibitem{bakhtiari2014online}
Ali~Shojaee Bakhtiari and Nizar Bouguila.
\newblock Online learning for two novel latent topic models.
\newblock In {\em Information and Communication Technology}, pages 286--295.
  Springer, 2014.

\bibitem{blei2006correlated}
David Blei and John Lafferty.
\newblock Correlated topic models.
\newblock {\em Advances in neural information processing systems}, 18:147,
  2006.

\bibitem{blei2003latent}
David~M Blei, Andrew~Y Ng, and Michael~I Jordan.
\newblock Latent dirichlet allocation.
\newblock {\em the Journal of machine Learning research}, 3:993--1022, 2003.

\bibitem{chen2013scalable}
Jianfei Chen, Jun Zhu, Zi~Wang, Xun Zheng, and Bo~Zhang.
\newblock Scalable inference for logistic-normal topic models.
\newblock In {\em Advances in Neural Information Processing Systems}, pages
  2445--2453, 2013.

\bibitem{favaro2011class}
Stefano Favaro, Georgia Hadjicharalambous, and Igor Pr{\"u}nster.
\newblock On a class of distributions on the simplex.
\newblock {\em Journal of Statistical Planning and Inference},
  141(9):2987--3004, 2011.

\bibitem{griffiths2004finding}
Thomas~L Griffiths and Mark Steyvers.
\newblock Finding scientific topics.
\newblock {\em Proceedings of the National Academy of Sciences}, 101(suppl
  1):5228--5235, 2004.

\bibitem{huang2016discovery}
Furong Huang.
\newblock Discovery of latent factors in high-dimensional data using tensor
  methods.
\newblock {\em arXiv preprint arXiv:1606.03212}, 2016.

\bibitem{klenke2014infinitely}
Achim Klenke.
\newblock Infinitely divisible distributions.
\newblock In {\em Probability Theory}, pages 331--349. Springer, 2014.

\bibitem{kolossiatis2011modeling}
Michalis Kolossiatis, Jim~E Griffin, and Mark~FJ Steel.
\newblock Modeling overdispersion with the normalized tempered stable
  distribution.
\newblock {\em Computational Statistics \& Data Analysis}, 55(7):2288--2301,
  2011.

\bibitem{Lichman:2013}
M.~Lichman.
\newblock {UCI} machine learning repository, 2013.

\bibitem{lijoi2005hierarchical}
Antonio Lijoi, Rams{\'e}s~H Mena, and Igor Pr{\"u}nster.
\newblock Hierarchical mixture modeling with normalized inverse-gaussian
  priors.
\newblock {\em Journal of the American Statistical Association},
  100(472):1278--1291, 2005.

\bibitem{lijoi2010models}
Antonio Lijoi and Igor Pr{\"u}nster.
\newblock Models beyond the dirichlet process.
\newblock {\em Bayesian nonparametrics}, 28:80, 2010.

\bibitem{mangili2015new}
Francesca Mangili and Alessio Benavoli.
\newblock New prior near-ignorance models on the simplex.
\newblock {\em International Journal of Approximate Reasoning}, 56:278--306,
  2015.

\bibitem{mimno2008gibbs}
David Mimno, Hanna~M Wallach, and Andrew McCallum.
\newblock Gibbs sampling for logistic normal topic models with graph-based
  priors.
\newblock 2008.

\bibitem{passos2011correlations}
Alexandre Passos, Hanna~M Wallach, and Andrew McCallum.
\newblock Correlations and anticorrelations in lda inference.
\newblock 2011.

\bibitem{sato2010topic}
Issei Sato and Hiroshi Nakagawa.
\newblock Topic models with power-law using pitman-yor process.
\newblock In {\em Proceedings of the 16th ACM SIGKDD international conference
  on Knowledge discovery and data mining}, pages 673--682. ACM, 2010.

\bibitem{tung2014spectral}
Hsiao-Yu Tung and Alex~J Smola.
\newblock Spectral methods for indian buffet process inference.
\newblock In {\em Advances in Neural Information Processing Systems}, pages
  1484--1492, 2014.

\end{thebibliography}
\end{document}